\documentclass[pdflatex,sn-mathphys-num]{sn-jnl}%

\usepackage{graphicx}%
\usepackage{multirow}%
\usepackage{amsmath,amssymb,amsfonts}%
\usepackage{amsthm}%
\usepackage{mathrsfs}%
\usepackage{enumitem}%
\usepackage[title]{appendix}%
\usepackage{xcolor}%
\usepackage{textcomp}%
\usepackage{manyfoot}%
\usepackage{booktabs}%
\usepackage{algorithm}%
\usepackage{algorithmicx}%
\usepackage{algpseudocode}%

\DeclareMathOperator*{\argmin}{arg\,min}
\newcommand{\E}{\mathbb{E}}
\newcommand{\R}{\mathbb{R}}
\newcommand{\X}{\mathcal{X}}
\newcommand{\CVaR}{\mathrm{CVaR}}
\newcommand{\VaR}{\mathrm{VaR}}
\newcommand{\Wass}{\mathsf{W}_1}
\newcommand{\orcidtext}{{\footnotesize(ORCID: 0000-0002-0121-6781)}}

\theoremstyle{thmstyleone}%
\newtheorem{theorem}{Theorem}%
\newtheorem{proposition}[theorem]{Proposition}%
\newtheorem{lemma}[theorem]{Lemma}%

\theoremstyle{thmstyletwo}%
\newtheorem{remark}{Remark}%

\theoremstyle{thmstylethree}%
\newtheorem{assumption}{Assumption}%

\begin{document}

\title[ACRS for Spectral Risk under Decision-Dependent Uncertainty]{Adaptive Conditional Residual Sampling for Spectral Risk Optimisation under Decision-Dependent Uncertainty}

\author*[1]{\fnm{Marcell T.} \sur{Kurbucz}~\href{https://orcid.org/0000-0002-0121-6781}{\orcidtext}}\email{m.kurbucz@ucl.ac.uk}

\affil*[1]{\orgdiv{Institute for Global Prosperity, The Bartlett}, \orgname{University College London}, \orgaddress{\street{9--11 Endsleigh Gardens}, \city{London}, \postcode{WC1H 0EH}, \country{United Kingdom}}}

\abstract{
\footnotesize
We consider minimising a spectral risk objective, the expected cost plus a weighted conditional value-at-risk, when the distribution of the uncertain parameters depends on the decision and can only be sampled through a stochastic simulator under a fixed budget of draws. We propose adaptive conditional residual sampling (ACRS). It takes one simulator draw at each point of a space-filling design, fits a heteroscedastic regression model of the decision-dependent distribution and reuses the standardised residual vectors as scenarios to construct a deterministic surrogate. ACRS searches this surrogate for several local optima, adds draws around the most promising ones, ranks the candidates on the simulator with common random numbers and finishes with sample-path local refinement using tail importance sampling. We show that the optimality gap of the surrogate minimiser is bounded by the Wasserstein error of the scenario distribution and that the refinement estimator is consistent. In the experiments, all methods share the budget accounting, the design sampler and the local refinement routine, and solutions are scored by normalised regret against reference optima. On 60 random instances of an analytic family with a closed-form objective, ACRS attains a mean regret of 11.8\% against 22.3--42.3\% for Bayesian optimisation, cross-entropy, kernel-density, stochastic-gradient and refinement-only baselines; its lead is largest at small budgets and, relative to the kernel-density method, tends to grow with the dimension. ACRS also has the lowest mean regret on two six-dimensional benchmarks and three development problems, and performs comparably to pure local refinement on three held-out problems frozen before testing.}


\maketitle

\section{Introduction}\label{sec:intro}

In many decision problems under uncertainty, the distribution of the uncertain parameters depends on the decision itself.
Such decision-dependent (or endogenous) uncertainty, whose modelling in stochastic programming was systematically developed by \citet{goel2006class}, arises, for example, when investment changes failure probabilities or the distribution of demand \cite{hellemo2018decision} or when prices change demand \cite{Bertsimas2020}.
A sample of the uncertain parameters drawn at one decision then misrepresents their distribution at another, whereas sample average approximation assumes a fixed sampling distribution \cite[Sect.~5.1]{Shapiro2021}, \cite{HomemDeMello2014}.
When, in addition, the decision maker controls exposure to extremes through the conditional value-at-risk (CVaR) \cite{Rockafellar2000,Rockafellar2002}, combined with the expected cost into a spectral risk objective \cite{Acerbi2002spec,Acerbi2002}, the optimiser has to track how the whole loss distribution, and in particular its upper tail, moves with the decision.

This paper studies that problem when the distribution $P_x$ of the uncertain parameters at decision $x$ is available only through a stochastic simulator, and every simulator draw is debited from a fixed budget $B$.
Three difficulties arise.
First, the objective at a single decision is a tail functional: an estimate of $\CVaR_{0.95}$ from $n$ draws rests on about $0.05\,n$ of them, so accurate pointwise evaluation is expensive.
Second, because $P_x$ moves with $x$, the objective is typically non-convex, with several basins whose depth depends on how the tail responds to the decision.
Third, the budget has to be split between learning about $P_x$ globally and optimising locally.
Response-surface methods such as Bayesian optimisation \cite{Jones1998,Huang2006,Frazier2018} model the objective directly and therefore spend many draws per evaluated decision to estimate the tail; local stochastic-approximation methods converge to a single local solution \cite{Hu2025} or, in performative prediction, to an equilibrium point \cite{cutler2024stochastic}; and conditional-distribution estimators that localise in decision space, through kernel distances or forest leaf co-membership \cite{Bertsimas2020,Athey2019,Cevid2022}, rely on the observations near a given decision, of which few lie in the tail when the dimension grows (Remark~\ref{rem:pooling}).

If the decision is regarded as a covariate, residual-based sample average approximation \cite{Kannan2025,Kannan2024} offers a way to pool information across all sampled decisions: a regression model captures how the location (and, in the heteroscedastic setting of \citet{Kannan2025}, the scale) of the uncertain parameters moves with the covariate, and the standardised residuals of all observations are reused as the scenario set at every covariate value.
This idea was developed for offline data with exogenous covariates and has recently been extended, for additive error models, to decision-dependent distributions \cite{Zhu2024,SunYu2026}, still with a given historical data set.
Simulation optimisation offers more freedom: the optimiser chooses where to sample, can reuse the simulator's random input (common random numbers) and can stop learning once the remaining budget is better spent on local search.

We propose adaptive conditional residual sampling (ACRS), a simulation-optimisation method built on this observation.
ACRS takes one simulator draw at each point of a space-filling design, fits a heteroscedastic residual model of $P_x$, and evaluates the spectral risk on the resulting scenario set.
The surrogate is deterministic and piecewise smooth, so it can be searched by multistart local optimisation without additional simulator draws.
Because the model is in general misspecified (for example, it cannot represent a tail index that changes with the decision), ACRS uses the surrogate only to propose candidates: it densifies the design around the best of them, ranks them on the simulator with common random numbers, and hands the winner to a sample-path local refinement with re-simulated finite differences and tail importance sampling.

\vspace{0.5em}
\noindent
The contributions are as follows:

\begin{enumerate}
\item ACRS transfers residual-based conditional modelling from offline contextual optimisation to budget-constrained, adaptive simulation optimisation of spectral risk under decision-dependent uncertainty.
\item We show that the optimality gap of any minimiser of the surrogate is at most $2(1+\lambda/(1-\alpha))L_C$ times the uniform Wasserstein error of the scenario distribution, where $L_C$ is the Lipschitz constant of the cost; that the defensive tail importance-sampling estimator used in the refinement has decision-independent, bounded weights and is consistent; and that the sample-path objective with re-simulated inputs converges uniformly.
\item We compare six methods under the same budget accounting, with a shared design sampler, a shared local refinement routine and a refinement-only control, on four benchmark families, scoring solutions by normalised regret against reference optima and evaluating on held-out problems that were frozen before any method was run on them.
\item ACRS has the lowest mean regret in all four families, although on the held-out problems its mean regret is only comparable to that of pure local refinement. Its advantage is largest at small budgets and, against kernel-based conditional modelling, tends to grow with the dimension. In two dimensions a kernel-based method is almost as good, and on two held-out problems at the smallest budget pure local refinement is better.
\end{enumerate}

The remainder of the paper is organised as follows.
Section~\ref{sec:related} reviews related work, Section~\ref{sec:problem} states the problem, Section~\ref{sec:method} describes ACRS and Section~\ref{sec:theory} its theoretical properties.
Sections~\ref{sec:benchmarks} and \ref{sec:protocol} present the benchmarks and the protocol, Section~\ref{sec:results} reports the results, Section~\ref{sec:discussion} discusses limitations, and Section~\ref{sec:conclusions} concludes the paper.

\section{Related work}\label{sec:related}

In stochastic programming, decision-dependent uncertainty is usually modelled through decisions that change the probabilities of a finite scenario set or reveal information \cite{goel2006class,apap2017models,hellemo2018decision}; robust linear optimisation with decision-dependent uncertainty sets was studied by \citet{Nohadani2018}. In machine learning and stochastic approximation, the performative-prediction literature \cite{Perdomo2020} and its optimisation counterparts \cite{Drusvyatskiy2023,cutler2024stochastic} analyse repeated retraining and stochastic gradient methods when the data distribution responds to the deployed decision, typically assuming smoothness and strong convexity, using risk-neutral objectives, and establishing convergence to equilibrium (performatively stable) points rather than to optima. We do not assume convexity, we target a tail-sensitive objective, and we treat the simulator as a black box except for control over its random input.

CVaR admits the variational representation of \citet{Rockafellar2000,Rockafellar2002}, and the combination of mean and CVaR used here is, up to a positive factor, a spectral risk measure in the sense of \citet{Acerbi2002spec}. \citet{Pichler2013} showed that spectral risk measures with a bounded spectrum are Lipschitz continuous in the Wasserstein distance, which we use to bound the error of the surrogate. Sensitivities of CVaR can be estimated by simulation \cite{HongLiu2009}, and \citet{Hu2025} developed a stochastic-approximation method for simulation optimisation of CVaR with three simulation calls per iteration. \citet{Cakmak2020} optimise risk measures by Bayesian optimisation when the environmental variable can be chosen at each query of the simulator, a form of control related to our use of common random numbers; their method models the simulator output as a Gaussian process in decision and environment jointly, and the distribution of the environmental variable does not depend on the decision. Importance sampling is a standard tool for tail estimation \cite{Glasserman2000,Glasserman2004}; defensive mixture proposals bound the second moment of the importance-weighted integrand relative to plain sampling \cite{Hesterberg1995,Owen2000}.

Efficient global optimisation \cite{Jones1998} and Gaussian-process methods for noisy simulators \cite{Huang2006,Ankenman2010} fit a Gaussian process to evaluations of the objective; reviews of Bayesian optimisation \cite{Shahriari2016,Frazier2018}, of derivative-free methods \cite{Rios2013} and of simulation optimisation \cite{Amaran2016,Fu2015} cover the broader field. The cross-entropy method \cite{Rubinstein2004,DeBoer2005} and differential evolution \cite{Storn1997,Price2005} are common population-based alternatives. These methods model or sample the objective; ACRS instead models the decision-dependent distribution and computes the objective from it.

Prescriptive analytics approximates the conditional distribution of uncertain parameters given covariates by reweighting observed samples, for example with kernels or random forests \cite{Bertsimas2020,Sadana2024}; generalised and distributional random forests provide such weights \cite{Athey2019,Cevid2022}. Residual-based sample average approximation fits a regression model and adds its (rescaled) residuals to the prediction at the new covariate \cite{Kannan2025}; distributionally robust variants \cite{Kannan2024} and extensions to decision-dependent distributions \cite{Zhu2024,SunYu2026} provide consistency guarantees. \citet{Bertsimas2020} also treat decisions that affect the uncertainty, by reweighting historical observations. These methods use historical data rather than observations collected adaptively by the optimiser. ACRS uses residual-based modelling within an adaptive simulation loop, followed by simulator-based candidate selection and local refinement.

With common random numbers, the sample-path objective is a deterministic function of the decision that can be optimised by deterministic methods \cite{Robinson1996,Kim2015}; variance reduction by common random numbers and antithetic variates is classical \cite{LEcuyer1994,Glasserman2004}. Our shared refinement follows this approach with bound-constrained L-BFGS-B \cite{Byrd1995}.

\section{Problem formulation}\label{sec:problem}

Let $\X\subset\R^d$ be a compact feasible set. For each decision $x\in\X$, the uncertain parameter vector $W\in\R^m$ follows a distribution $P_x$ that may depend on $x$. We assume that $P_x$ is accessible only through a simulator, also called the oracle, with a controllable random input.

\begin{assumption}[Controllable simulator]\label{ass:sim}
There is a measurable map $T:\X\times\R^{d_Z}\to\R^m$ and a known input distribution $\gamma=\mathcal{N}(0,I_{d_Z})$ such that $T(x,Z)\sim P_x$ for $Z\sim\gamma$. The optimiser may evaluate $T(x,z)$ at inputs $z$ of its choice; each evaluation costs one unit of a total budget $B$.
\end{assumption}

Many simulators driven by a seedable random number generator can be written in this form, for example by mapping a fixed number of uniforms through inverse transforms, and the assumption makes common random numbers and changes of measure in the input space available to every method. Given a measurable cost $C:\R^m\times\X\to\R$, the loss at decision $x$ is $L(x,Z)=C(T(x,Z),x)$, with distribution function $F_x$. For $\lambda\ge 0$ and $\alpha\in(0,1)$ we minimise
\begin{equation}
  J(x) \;=\; \E_{W\sim P_x}\bigl[C(W,x)\bigr] + \lambda\,\CVaR_\alpha^{P_x}\bigl(C(W,x)\bigr),
  \qquad x^\star\in\argmin_{x\in\X} J(x),
  \label{eq:objective}
\end{equation}
where the superscript stresses that both the expectation and the CVaR are taken under the decision-dependent distribution. Larger values of $C$ are losses, and
\begin{equation}
  \CVaR_\alpha(L) = \inf_{\tau\in\R}\Bigl\{\tau+\tfrac{1}{1-\alpha}\,\E\bigl[(L-\tau)_+\bigr]\Bigr\}
  \label{eq:cvar}
\end{equation}
is the Rockafellar--Uryasev representation \cite[Thm~1]{Rockafellar2000}, which \citet[Thm~10]{Rockafellar2002} extended to general loss distributions; the minimisers form a closed interval whose left endpoint is the $\alpha$-quantile $\VaR_\alpha(L)$. Equivalently, $J(x)=\rho(F_x)$ with the spectral risk measure
\begin{equation}
  \rho(F) = \int_0^1 F^{-1}(u)\,\phi(u)\,\mathrm{d}u,\qquad
  \phi(u) = 1+\frac{\lambda}{1-\alpha}\,\mathbf{1}\{u\ge\alpha\},
  \label{eq:spectral}
\end{equation}
whose non-decreasing spectrum $\phi$ gives unit weight to the mean and additional weight $\lambda$ to the upper $\alpha$-tail; up to the normalising constant $\int_0^1\phi=1+\lambda$, $\rho$ is a spectral risk measure. In the profit convention of \citet[Def.~3.1, Thm~4.1]{Acerbi2002spec}, $\rho(F_L)=(1+\lambda)M_\psi(-L)$ with the admissible spectrum $\psi(p)=\phi(1-p)/(1+\lambda)$; the identity $\CVaR_\alpha(L)=(1-\alpha)^{-1}\int_\alpha^1F_L^{-1}(u)\,\mathrm{d}u$, valid for general distributions \cite[Prop.~3.2]{Acerbi2002}, links \eqref{eq:spectral} to \eqref{eq:objective}. We write $\bar\phi=\sup_u\phi(u)=1+\lambda/(1-\alpha)$. Throughout the experiments $\lambda=0.7$ and $\alpha=0.95$ unless stated otherwise.

Given simulated losses $\ell_1,\ldots,\ell_n$ with normalised weights $\bar w_i\ge 0$, $\sum_i\bar w_i=1$, the objective is estimated by the spectral risk of the weighted empirical distribution $\hat F=\sum_i\bar w_i\delta_{\ell_i}$,
\begin{equation}
  \rho(\hat F) = \sum_{i}\bar w_i\ell_i + \lambda\Bigl\{\hat v+\frac{1}{1-\alpha}\sum_i\bar w_i(\ell_i-\hat v)_+\Bigr\},
  \label{eq:estimator}
\end{equation}
where $\hat v$ is the $\alpha$-quantile of $\hat F$. For equal weights we use the common tail-average estimator of CVaR, namely the mean of the losses at or above the empirical $\alpha$-quantile computed by linear interpolation between order statistics. For losses with a finite first moment whose distribution is continuous at its $\alpha$-quantile, which holds in all our benchmarks, it differs from \eqref{eq:estimator} by a term that vanishes almost surely as $n\to\infty$. Every method uses the tail-average estimator for equally weighted samples and \eqref{eq:estimator} for importance-weighted samples, and all Monte Carlo final evaluations use the tail-average estimator. The task is to return a decision $\hat x$ with small $J(\hat x)-J(x^\star)$ after at most $B$ simulator draws.

\section{Adaptive conditional residual sampling}\label{sec:method}

\begin{figure}[t]
\centering
\includegraphics[width=0.75\linewidth]{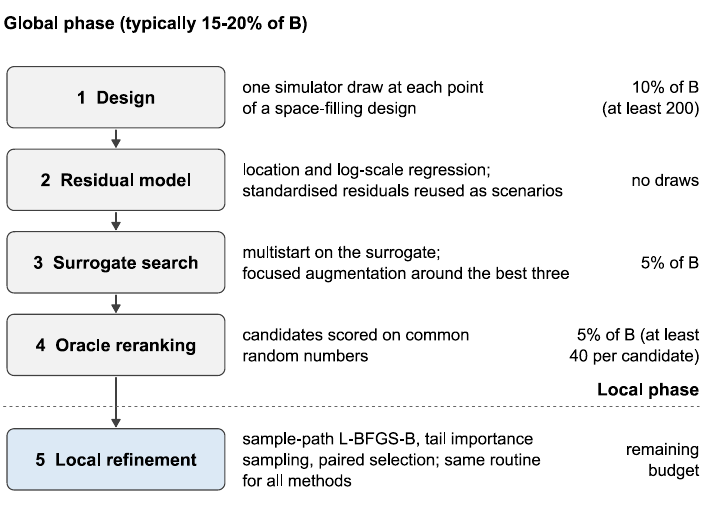}
\caption{Schematic of ACRS. Phases 1--4 form the global phase; phase 5 is the local refinement routine, which all methods in the experiments use whenever enough budget remains. Budget shares refer to the total budget $B$ of simulator draws; the reranking budget is spent only if the surrogate has more than one distinct candidate. Every simulator draw is debited from one counter with hard cap $B$; on exhaustion the incumbent is returned}
\label{fig:schematic}
\end{figure}

ACRS has a global phase that learns $P_x$ and locates candidate basins, and a local phase that refines the selected candidate on the simulator (Fig.~\ref{fig:schematic}, Algorithm~\ref{alg:acrs}). The global phase typically uses 15--20\% of the budget (at most 31\% when many candidates are reranked at the smallest budget); the local phase, which uses the same routine as all competing methods in our experiments, uses the rest.

\subsection{Space-filling design with single draws}\label{sec:design}

ACRS draws a Latin hypercube design \cite{McKay1979} of $n_0=\max(200,\lfloor 0.1B\rceil)$ points ($\lfloor\cdot\rceil$ denotes rounding to the nearest integer) $x_1,\ldots,x_{n_0}$ over $\X$ and takes one simulator draw $W_i=T(x_i,Z_i)$, $Z_i\sim\gamma$, at each. Taking one draw per point allows the design to cover more distinct decisions for a fixed number of draws; instead of replicating at individual points, the residual model below pools information across the design. When $\X$ is not a box, the design must cover the feasible region uniformly; for the polytope of Section~\ref{sec:dgp} we map Latin hypercube uniforms to the simplex by sorted spacings and reject points that violate the remaining bounds (Appendix~\ref{app:dgp}), and for the budget constraint of problem E2 we use uniform rejection sampling.

\subsection{Heteroscedastic residual model}\label{sec:model}

Let $g:\R^m\to\R^m$ act componentwise, with $g_j(w)=\log w$ if all observed $W_{ij}$ are positive and $g_j(w)=w$ otherwise, and let $y_i=g(W_i)$. The location model is a ridge regression on quadratic and radial-basis features,
\begin{equation}
  \hat f(x)=\hat\Theta^\top\psi(x),\qquad
  \psi(x)=\Bigl(1,\;x,\;\{x_kx_l\}_{k\le l},\;\bigl\{\exp\bigl(-\|x-c_r\|^2/(2\omega^2)\bigr)\bigr\}_{r=1}^{R}\Bigr),
  \label{eq:location}
\end{equation}
where the $R=\max\{5,\min(60,\lfloor N/20\rfloor)\}$ centres $c_r$ are $k$-means centres of the $N$ design points and $\omega$ is the median distance from a centre to its nearest neighbouring centre. The ridge penalty is shared by all components and is chosen by generalised cross-validation \cite{Golub1979} on a logarithmic grid of 15 values between $10^{-6}$ and $10$ times the largest squared singular value of the feature matrix. With residuals $r_i=y_i-\hat f(x_i)$, the log-scale of each component is linear in $x$,
\begin{equation}
  \log\bigl(r_{ij}^2+c_j\bigr)\approx a_{j0}+a_j^\top x_i,\qquad
  \hat s_j(x)=\mathrm{clip}\Bigl(\exp\bigl\{(\hat a_{j0}+\hat a_j^\top x)/2\bigr\},\;\underline{s}_j,\;\overline{s}_j\Bigr),
  \label{eq:scale}
\end{equation}
fitted by least squares with the stabilising constant $c_j=0.05\,\mathrm{median}_i\,r_{ij}^2$ and clipped to the range $[\underline s_j,\overline s_j]$ of the fitted values on the design. The standardised residual vectors $e_i=r_i\oslash\hat s(x_i)$ ($\oslash$ denotes componentwise division) are kept jointly, so the dependence between the components of $W$ is preserved. The scenario set at any decision $x$ is
\begin{equation}
  \hat W_i(x) = g^{-1}\bigl(\hat f(x)+\hat s(x)\odot e_i\bigr),\qquad i=1,\ldots,N,
  \label{eq:scenarios}
\end{equation}
and the surrogate objective is the spectral risk of the empirical distribution of the scenario losses,
\begin{equation}
  \hat J_N(x)=\rho\Bigl(\tfrac{1}{N}\textstyle\sum_{i=1}^N\delta_{C(\hat W_i(x),\,x)}\Bigr).
  \label{eq:surrogate}
\end{equation}
In the implementation the CVaR part of \eqref{eq:surrogate} is computed with the tail-average estimator of Section~\ref{sec:problem}, which is asymptotically equivalent to \eqref{eq:surrogate} in the sense stated there. Because the same $\hat s$ standardises the residuals and rescales them in \eqref{eq:scenarios}, a bias of the log-scale fit \eqref{eq:scale} that does not depend on $x$ cancels; the stabilising constant $c_j$ adds a small $x$-dependent bias where the scale is small. The model is well specified when $g(W)=f(x)+s(x)\odot\zeta$ with $\zeta$ independent of $x$, and misspecified otherwise, for example when the tail index or the correlation of $W$ depends on $x$. Two properties are important for optimisation: $\hat J_N$ is a deterministic, piecewise smooth function of $x$, and at every decision its $\alpha$-tail is represented by $(1-\alpha)N$ scenarios, independently of the dimension and of how many design points lie near $x$ (Remark~\ref{rem:pooling}).

\subsection{Surrogate multistart and focused augmentation}\label{sec:search}

The surrogate is evaluated at 200 fresh space-filling points (no simulator draws are needed), and L-BFGS-B \cite{Byrd1995} with numerical gradients (step $10^{-3}$, at most 100 iterations) is started from the 20 best. The end points are sorted by surrogate value and kept if their sup-norm distance to every kept point exceeds $0.05$; at most $k_{\max}=8$ candidates are retained. ACRS then spends $n_a=\lfloor 0.05B\rceil$ draws on focused augmentation: point $i$ is drawn uniformly from the box of half-width $0.1$ (relative to the range of each coordinate) around the $((i-1)\bmod k'+1)$-th best candidate, $k'=\min(3,k)$, and projected onto $\X$. The model is refitted on all $N=n_0+n_a$ points and the candidate list is recomputed. Augmentation improves the model where the decision is likely to be made, at the price of a less uniform design.

\subsection{Oracle reranking with common random numbers}\label{sec:rerank}

If $k>1$ candidates remain, each is evaluated on the simulator with the same $n_r=\max\{40,\lfloor 0.05B/k\rfloor\}$ antithetic inputs $(z,-z)$, and the candidates are ordered by the estimate \eqref{eq:estimator}. Because all candidates share the inputs, their differences, which determine the ranking, are estimated with common random numbers. Reranking corrects the order in which the surrogate, possibly biased in the tail, ranks the basins; the ordered list is passed to the local refinement.

\subsection{Shared local refinement}\label{sec:refine}

The refinement receives an ordered list of starting points and is the same for every method in our experiments. It proceeds in three steps. First, a pilot sample of $n_s$ antithetic inputs is simulated at the first start; the shift $\hat\theta$ is the mean of the inputs whose losses are at or above the empirical $\alpha$-quantile, rescaled to norm at most $2.5$. All later samples are drawn from the defensive mixture \cite{Hesterberg1995}
\begin{equation}
  q=(1-\beta)\,\mathcal{N}(0,I_{d_Z})+\beta\,\mathcal{N}(\hat\theta,I_{d_Z}),\qquad
  w(z)=\Bigl[(1-\beta)+\beta\exp\bigl(\hat\theta^\top z-\tfrac12\|\hat\theta\|^2\bigr)\Bigr]^{-1},
  \label{eq:is}
\end{equation}
with $\beta=0.5$, and the objective is estimated by \eqref{eq:estimator} with self-normalised weights $\bar w_i=w(z_i)/\sum_j w(z_j)$.

Second, for each start in turn, a fresh sample $z_1,\ldots,z_{n_\ell}$ from $q$ is drawn and kept fixed, which turns $x\mapsto\rho(\hat F^w_{n_\ell,x})$ into a deterministic sample-path objective. L-BFGS-B (at most 20 iterations) minimises it with forward-difference gradients of step $h=0.005$ that re-simulate $T(x+he_j,z_i)$ with the same inputs, so the gradient accounts for the dependence of $P_x$ on $x$ (Remark~\ref{rem:resim}). The best point visited is retained, which protects against non-smooth sample paths.

Third, a selection sample of $n_s$ inputs from $q$ is drawn once, before the local searches; every start and every end point is simulated on these common inputs, at a cost of $n_s$ draws each, and the best score determines the returned decision. A new start is refined only if at least $2n_s+n_\ell(d+1)$ draws remain, and $n_s$ draws are reserved while a run is in progress so that its end point can always be scored. If the budget is exhausted inside a run, the best point visited so far is scored with the reserved draws and the incumbent is returned.

\subsection{Algorithm and cost}

Algorithm~\ref{alg:acrs} summarises ACRS. The budget-dependent sample sizes are listed in Appendix~\ref{app:settings}; for $B\le 10{,}000$ they are $n_\ell=80$ and $n_s=150$. The non-simulator work consists of fitting the model, which costs one singular value decomposition of an $N\times p$ feature matrix with $p\le 1+d+d(d+1)/2+60$, and of evaluating the surrogate, which costs $O(N(m+p))$ operations per decision. In our experiments the median overhead was 1--9\,s per run (Section~\ref{sec:cost}).

\begin{algorithm}[t]
\caption{Adaptive conditional residual sampling (ACRS)}\label{alg:acrs}
\begin{algorithmic}[1]
\Require simulator $T$, cost $C$, feasible set $\X$, budget $B$, risk parameters $(\lambda,\alpha)$
\Ensure decision $\hat x\in\X$
\State $n_0\gets\max(200,\lfloor 0.1B\rceil)$; draw a space-filling design $x_1,\ldots,x_{n_0}$; $W_i\gets T(x_i,Z_i)$ \Comment{one draw per point}
\State fit the residual model \eqref{eq:location}--\eqref{eq:scale}; form $\hat J_N$ by \eqref{eq:surrogate}
\State $\mathcal{C}\gets$ distinct local minimisers of $\hat J_N$ from 20 L-BFGS-B starts, at most 8 \Comment{no draws}
\State draw $\lfloor 0.05B\rceil$ points around the 3 best elements of $\mathcal{C}$; one draw each; refit; recompute $\mathcal{C}$
\If{$|\mathcal{C}|>1$}
  \State score all $x\in\mathcal{C}$ on common antithetic inputs ($\max\{40,\lfloor 0.05B/|\mathcal{C}|\rfloor\}$ each); sort $\mathcal{C}$
\EndIf
\State estimate the tail shift $\hat\theta$ at the first element of $\mathcal{C}$; draw a selection sample from $q$ in \eqref{eq:is}
\For{each $x_0\in\mathcal{C}$ while at least $2n_s+n_\ell(d+1)$ draws remain}
  \State run L-BFGS-B on the weighted sample-path objective from $x_0$ with re-simulated differences
  \State score $x_0$ and the best point visited on the selection sample
\EndFor
\State \Return the point with the best selection score (incumbent if the budget is exhausted)
\end{algorithmic}
\end{algorithm}

\section{Theoretical properties}\label{sec:theory}

We establish three properties of ACRS: a bound on the optimality gap of the surrogate minimiser, consistency of the importance-sampling estimator, and convergence of the sample-path objective. We do not derive convergence rates for the whole method; the budget is finite and the surrogate is misspecified in general, which is why the final decision is selected using simulator evaluations.

\begin{assumption}[Lipschitz cost]\label{ass:lip}
There is $L_C<\infty$ such that $|C(w,x)-C(w',x)|\le L_C\|w-w'\|$ for all $w,w'\in\R^m$ and $x\in\X$, and $\E\|T(x,Z)\|<\infty$ for every $x\in\X$.
\end{assumption}

\begin{lemma}[Wasserstein continuity]\label{lem:wass}
For distribution functions $F,G$ on $\R$ with finite first moments, $|\rho(F)-\rho(G)|\le\bar\phi\,\Wass(F,G)$, where $\Wass$ is the $1$-Wasserstein distance.
\end{lemma}

\begin{proof}
On the real line $\Wass(F,G)=\int_0^1|F^{-1}(u)-G^{-1}(u)|\,\mathrm{d}u$ \cite{Vallender1974}. Hence $|\rho(F)-\rho(G)|\le\int_0^1|F^{-1}(u)-G^{-1}(u)|\,\phi(u)\,\mathrm{d}u\le\bar\phi\,\Wass(F,G)$. This is a special case of \citet[Cor.~11]{Pichler2013}.
\end{proof}

\begin{proposition}[Optimality gap of the surrogate minimiser]\label{prop:surrogate}
Let Assumption~\ref{ass:lip} hold, let $\hat P_x$ be any probability distribution on $\R^m$ with a finite first moment, for example the scenario distribution \eqref{eq:scenarios}, and let $\hat J(x)$ be the spectral risk of $C(\hat W,x)$ with $\hat W\sim\hat P_x$. Define $\varepsilon=\sup_{x\in\X}\Wass(\hat P_x,P_x)$, with $\Wass$ taken with respect to the norm of Assumption~\ref{ass:lip}. Then
\begin{enumerate}
\item[(a)] $\sup_{x\in\X}|\hat J(x)-J(x)|\le\bar\phi L_C\,\varepsilon$;
\item[(b)] every $\eta$-minimiser $\hat x$ of $\hat J$, that is, $\hat J(\hat x)\le\inf_{\X}\hat J+\eta$, satisfies
\begin{equation}
  J(\hat x)-\inf_{\X}J\;\le\;2\Bigl(1+\frac{\lambda}{1-\alpha}\Bigr)L_C\,\varepsilon+\eta .
  \label{eq:gap}
\end{equation}
\end{enumerate}
\end{proposition}

\begin{proof}
(a) For a coupling $\pi$ of $\hat P_x$ and $P_x$, the pair $(C(\hat W,x),C(W,x))$ with $(\hat W,W)\sim\pi$ is a coupling of the two loss distributions, so their Wasserstein distance is at most $\E_\pi|C(\hat W,x)-C(W,x)|\le L_C\,\E_\pi\|\hat W-W\|$. Taking the infimum over couplings and applying Lemma~\ref{lem:wass} gives $|\hat J(x)-J(x)|\le\bar\phi L_C\Wass(\hat P_x,P_x)\le\bar\phi L_C\varepsilon$. (b) Let $\delta=\bar\phi L_C\varepsilon$. For every $x'\in\X$, $J(\hat x)\le\hat J(\hat x)+\delta\le\hat J(x')+\eta+\delta\le J(x')+\eta+2\delta$; taking the infimum over $x'$ proves \eqref{eq:gap}.
\end{proof}

The bound applies to the spectral risk of the empirical scenario distribution defined in \eqref{eq:surrogate}; the implementation uses the asymptotically equivalent tail-average estimator described in Section~\ref{sec:model}.

\begin{remark}[Consistency of the surrogate]\label{rem:consistency}
Suppose that $g(W)=f(x)+s(x)\odot\zeta$ with $\zeta$ independent of $x$ (the location--scale case), that $\hat f\to f$ and $\hat s\to s$ uniformly on $\X$, and that the empirical distribution of the estimated standardised residuals converges to that of $\zeta$ in $\Wass$. The uniform consistency of $\hat f$ and $\hat s$ is an assumption: it requires, among other things, that $f$ lies in the span of the features of \eqref{eq:location} and that the stabilising constants $c_j$ in \eqref{eq:scale} vanish asymptotically, because for $c_j>0$ the log-scale fit carries a bias that depends on $x$. If $g$ is the identity, these conditions imply $\varepsilon\to 0$ as $N\to\infty$, so that the surrogate minimiser is asymptotically optimal by \eqref{eq:gap}; with a logarithmic transform, the same conclusion additionally requires the scenario distributions $\hat P_x$ to be uniformly integrable as $N\to\infty$, because the exponential is not Lipschitz. Related conditions are analysed by \citet{Kannan2025,Kannan2024} for exogenous covariates and, for location-only models with decision-dependent distributions, by \citet{Zhu2024,SunYu2026}. Under misspecification, $\varepsilon$ need not vanish, and \eqref{eq:gap} bounds the error from using the surrogate alone. Assumption~\ref{ass:lip} holds for the analytic family and for E1--E6, whose costs are piecewise linear in $W$, but not for DGP1 and DGP2, whose costs grow superlinearly in $W$; for those the proposition applies only on bounded sets of scenarios. The bound is uniform in $x$ and therefore pessimistic; for ACRS it suffices that the surrogate places a candidate in the basin of a global minimiser, after which the oracle phases take over. Section~\ref{sec:phases} quantifies both effects.
\end{remark}

\begin{remark}[Pooling and the tail]\label{rem:pooling}
In \eqref{eq:surrogate} the upper $\alpha$-tail at every decision is formed by $(1-\alpha)N$ scenarios. A conditional estimator that reweights the design points by their distance to $x$, such as a kernel estimator with bandwidth $b$, has an effective sample size of order $Nb^d$ and therefore about $(1-\alpha)Nb^d$ tail observations at $x$. Pooling trades the variance of the local estimate for the bias of a misspecified global model; in our experiments this trade-off increasingly favours pooling as the dimension grows and the budget shrinks (Section~\ref{sec:res-analytic}).
\end{remark}

\begin{proposition}[Defensive tail importance sampling]\label{prop:is}
Let $q$ and $w$ be as in \eqref{eq:is} with $\beta\in[0,1)$ and a fixed shift $\theta$. Then
\begin{enumerate}
\item[(a)] $0<w(z)\le(1-\beta)^{-1}$ for all $z$, and $w$ does not depend on the decision $x$;
\item[(b)] for every measurable $h$ with $\E_\gamma h(Z)^2<\infty$, $\E_q\bigl[w(Z)^2h(Z)^2\bigr]\le(1-\beta)^{-1}\E_\gamma\bigl[h(Z)^2\bigr]$;
\item[(c)] if $Z_1,Z_2,\ldots$ are independent with distribution $q$ and $\E_\gamma|L(x,Z)|<\infty$, the weighted empirical distribution $\hat F^w_{n,x}=\sum_{i\le n}\bar w_i\delta_{L(x,Z_i)}$ satisfies $\Wass(\hat F^w_{n,x},F_x)\to 0$ almost surely, and hence $\rho(\hat F^w_{n,x})\to J(x)$ almost surely.
\end{enumerate}
\end{proposition}

\begin{proof}
(a) $w=\mathrm{d}\gamma/\mathrm{d}q$ and the exponential term is positive. The change of measure acts on the simulator input only, so $w$ is the same for every decision. (b) $\E_q[w^2h^2]=\E_\gamma[wh^2]\le\sup_z w(z)\,\E_\gamma[h^2]$. (c) By the strong law of large numbers, $n^{-1}\sum_i w(Z_i)\to 1$ and $n^{-1}\sum_i w(Z_i)\,k(L(x,Z_i))\to\E_\gamma k(L(x,Z))$ almost surely for every $k$ in a countable convergence-determining class of bounded continuous functions and for $k(\ell)=|\ell|$. Hence $\hat F^w_{n,x}$ converges weakly to $F_x$ together with its first absolute moment, which is equivalent to convergence in $\Wass$ \cite[Chap.~6]{Villani2009}. Lemma~\ref{lem:wass} completes the proof.
\end{proof}

Part (a) makes importance sampling compatible with sample-path optimisation: one weighted sample serves every decision visited by the local run, so differences between decisions remain paired. Part (b) gives the defensive bound: at $\beta=0.5$, the second moment of the weighted integrand $w(Z)h(Z)$ is at most twice that of $h(Z)$ under plain sampling, for any $h$ and regardless of the estimated shift; see \citet[Eq.~(8)]{Owen2000} for this bound and \citet[Sect.~6.1]{Hesterberg1995} for the defensive mixture. For self-normalised estimators of expectations, \citet[Sect.~6.1]{Hesterberg1995} shows the same factor for the asymptotic variance; for the estimator of $J$, a nonlinear functional of $\hat F^w_{n,x}$, we do not claim a variance bound. Conditional on the pilot sample, the refinement samples are independent of $\hat\theta$, so Proposition~\ref{prop:is} applies conditionally.

\begin{proposition}[Sample-path refinement]\label{prop:samplepath}
In addition to the conditions of Proposition~\ref{prop:is}, suppose that $x\mapsto L(x,z)$ is continuous on $\X$ for $\gamma$-almost every $z$ and that $\E_\gamma\sup_{x\in\X}|L(x,Z)|<\infty$. Let $\hat J_n(x)=\rho(\hat F^w_{n,x})$ be the sample-path objective built from one fixed weighted sample. Then $\sup_{x\in\X}|\hat J_n(x)-J(x)|\to 0$ almost surely. Consequently, if $\hat x_n$ is an $\eta_n$-minimiser of $\hat J_n$ with $\eta_n\to 0$, then $J(\hat x_n)\to\inf_\X J$ almost surely, and every limit point of $(\hat x_n)$ minimises $J$.
\end{proposition}

\begin{proof}[Proof (sketch)]
By \eqref{eq:cvar}, $\hat J_n(x)=\min_\tau\hat\psi_n(x,\tau)$ and $J(x)=\min_\tau\psi(x,\tau)$ with $\hat\psi_n(x,\tau)=\sum_i\bar w_i\{L(x,Z_i)+\lambda\tau+\lambda(L(x,Z_i)-\tau)_+/(1-\alpha)\}$ and $\psi$ its expectation under $\gamma$. Let $\bar L=\E_\gamma\sup_{x\in\X}|L(x,Z)|$. By Markov's inequality, every $\alpha$-quantile $v$ of $F_x$ satisfies $|v|\le\bar L/\min(\alpha,1-\alpha)$, since otherwise $\Pr(L\ge v)<1-\alpha$ or $\Pr(L\le v)<\alpha$. The same argument applied to $\hat F^w_{n,x}$, whose weighted mean of $\sup_x|L(x,Z_i)|$ converges to $\bar L$ almost surely, bounds its $\alpha$-quantiles by $2\bar L/\min(\alpha,1-\alpha)$ for all $x$ and all large $n$. Hence the minimisation over $\tau$ can be restricted to $[-M,M]$ with $M=2\bar L/\min(\alpha,1-\alpha)$. Write $\hat\psi_n$ as the ratio of $n^{-1}\sum_iw(Z_i)\{\cdots\}$ and $n^{-1}\sum_iw(Z_i)\to1$. The summands of the numerator are continuous in $(x,\tau)$ and dominated by $w(Z)(1+\lambda+\lambda/(1-\alpha))(\sup_x|L(x,Z)|+M)$, whose expectation under $q$ equals its expectation without the weight under $\gamma$ and is therefore finite; since $\gamma$ and $q$ are mutually absolutely continuous, continuity $\gamma$-almost surely is also continuity $q$-almost surely. The uniform law of large numbers \cite[Thm~9.60]{Shapiro2021}, applied to these sample averages under $q$, gives $\hat\psi_n\to\psi$ uniformly on $\X\times[-M,M]$, and uniform convergence passes to the minima over $\tau$. The consequences follow as in Proposition~\ref{prop:surrogate}(b) and by compactness of $\X$ and continuity of $J$; see also \citet[Thm~5.3]{Shapiro2021} and \citet[Thm~4.2]{Robinson1996}.
\end{proof}

Proposition~\ref{prop:samplepath} concerns near-minimisers of the sample-path objective. It does not imply that the finitely many L-BFGS-B iterations of the refinement reach such a minimiser, so it supports the estimator used by the refinement rather than the algorithm as a whole.

\begin{remark}[Re-simulated versus frozen differences]\label{rem:resim}
For fixed inputs, a difference quotient of the sample-path objective that re-simulates $T(x+he_j,z_i)$ approximates the pathwise derivative of $L(x,z_i)=C(T(x,z_i),x)$, which contains both $\partial_x C$ and $\nabla_w C\,\partial_x T$. A scheme that keeps the scenarios $W_i$ fixed and perturbs only the second argument of $C$ omits the second term, so its gradient estimate is biased in general when $P_x$ depends on $x$. When the sample path is discontinuous in $x$ (indicator-driven regime switches as in the jump variant of Section~\ref{sec:analytic} or the crash regime of problem E2), finite differences give only heuristic directions. Sample-path analysis covers non-smooth sample functions and approximate minimisers \cite[Thm~4.2]{Robinson1996}; keeping the best point visited and selecting by paired scores on common inputs are safeguards of the shared refinement.
\end{remark}

\section{Benchmark families}\label{sec:benchmarks}

We use four families (Table~\ref{tab:families}) that differ in how $P_x$ depends on $x$, in the dimension, and in their role in the development of the method. The objective is \eqref{eq:objective} with $\alpha=0.95$ throughout.

\begin{table}[t]
\caption{Benchmark families. $d$: decision dimension; $m$: dimension of $W$; $B$: simulation budget. Each analytic instance has one budget, drawn at random. Final evaluation is exact or uses a large sample of common random numbers shared by all methods within a replication}
\label{tab:families}
\begin{tabular*}{\textwidth}{@{\extracolsep\fill}lccccll@{}}
\toprule
Family & $d$ & $m$ & Units & $B$ (k) & Reps & Final evaluation \\
\midrule
Analytic family & 2--8 & 3 & 60 instances & 2, 5 or 10 & 4 & exact \\
DGP1, DGP2$^{a}$ & 6 & 5 & 18 settings & 5, 10, 30 & 30 & 20,000 draws \\
E1--E3 (development) & 3--5 & 2--5 & 9 settings & 2, 5, 10 & 20 & 100,000 draws \\
E4--E6 (held-out) & 3--5 & 3--5 & 9 settings & 2, 5, 10 & 20 & 100,000 draws \\
\botrule
\end{tabular*}
\footnotetext[a]{Each with $\lambda\in\{0.5,0.7,0.9\}$.}
\end{table}

\subsection{Analytic family with closed-form objective}\label{sec:analytic}

The analytic family allows exact evaluation and controlled variation of the dimension, the number of basins and the strength of decision dependence. For $x\in[0,1]^d$, let $W=\mu(x)+\sigma(x)\odot(\Lambda(x)Y)$, where $Y$ is a standard trivariate Student-$t$ vector with $\nu(x)$ degrees of freedom and $\Lambda(x)\Lambda(x)^\top$ is the equicorrelation matrix with parameter $\varrho(x)$. The mean has $K$ Gaussian wells,
\begin{equation*}
\mu(x)=2\cdot\mathbf{1}-A\bigl(h\odot\varphi(x)\bigr),\qquad \varphi_k(x)=\exp\bigl(-\|x-a_k\|^2/(2r^2)\bigr),\quad k=1,\ldots,K,
\end{equation*}
and the scale, the tail index and the correlation move with the decision, with strength controlled by $\delta\ge 0$:
\begin{align*}
\sigma(x)&=\exp\Bigl\{\log 0.3+\delta\,G(x-\tfrac12\mathbf{1})+\delta\textstyle\sum_k e_k\varphi_k(x)\Bigr\},\\
\nu(x)&=2.5+\frac{7.5}{1+\exp\{-4\delta\,u^\top(x-\frac12\mathbf{1})/\sqrt d\}},\\
\varrho(x)&=0.2+\frac{0.6}{1+\exp\{-4\delta\,v^\top(x-\frac12\mathbf{1})/\sqrt d\}}.
\end{align*}
The cost is $C(W,x)=0.3\|x-x_c\|^2+\mathbf{1}^\top W$. Since $\mathbf{1}^\top\sigma\odot(\Lambda Y)$ is a scaled Student-$t$ variable, the loss is $m_L(x)+s_L(x)T_{\nu(x)}$ with $m_L(x)=0.3\|x-x_c\|^2+\mathbf{1}^\top\mu(x)$, $s_L(x)^2=(1-\varrho)\|\sigma\|^2+\varrho(\mathbf{1}^\top\sigma)^2$, and
\begin{equation}
  J(x)=(1+\lambda)\,m_L(x)+\lambda\,s_L(x)\,\CVaR_\alpha(T_{\nu(x)}),\qquad
  \CVaR_\alpha(T_\nu)=\frac{\nu+t_\alpha^2}{\nu-1}\,\frac{f_\nu(t_\alpha)}{1-\alpha},
  \label{eq:analyticJ}
\end{equation}
where $t_\alpha$ and $f_\nu$ are the $\alpha$-quantile and the density of the standard $t_\nu$ distribution. In the jump variant, the first component of $W$ receives an additional jump of size $3s_L(x)$ with a decision-dependent probability between 0.02 and 0.08, which creates a second tail regime and discontinuous sample paths; $J$ remains available in closed form up to a scalar root (Appendix~\ref{app:analytic}). The simulator input is $Z\in\R^4$ ($\R^5$ with the jump): three components drive $Y$, one the $\chi^2$ mixing variable by inverse transform, and one the jump indicator. Every instance draws the well centres, depths and loadings, the vectors $u,v$, the matrix $G$ and the centre $x_c$ at random (Appendix~\ref{app:analytic}).

For the final test, 60 new instances were generated with $d\in\{2,4,6,8\}$, $K\in\{1,2,4\}$, $\delta\in\{0.3,1,2,3\}$ and $B\in\{2000,5000,10000\}$ drawn uniformly, with the jump variant selected with probability 0.3; the realised design has 16, 16, 11 and 17 instances with $d=2,4,6,8$, 25, 16 and 19 with $B=2000,5000,10000$, and 12 with jumps. None of these instances was used during development.

\subsection{DGP1 and DGP2}\label{sec:dgp}

These two six-dimensional benchmarks have a polytope feasible set. The decision $x=(x_1,\ldots,x_6)$ satisfies $x_j\in[0,0.7]$ for $j\le 5$, $\sum_{j\le 5}x_j\le 0.85$ and $x_6\in[0,1]$. DGP1 couples five Student-$t$ marginals with decision-dependent location, scale and degrees of freedom through a Gaussian copula with decision-dependent correlation; DGP2 uses log-normal marginals. The costs combine damage, overflow, exposure and delivery terms (DGP1) or holding, shortage, procurement and coordination terms (DGP2) with a quadratic decision cost. The complete specification is given in Appendix~\ref{app:dgp}. Design points are sampled uniformly from the feasible polytope (Section~\ref{sec:design}).

\subsection{External problems}\label{sec:external}

Six stylised operations problems with decision-dependent uncertainty that is not of location--scale form were specified independently of the method (Appendix~\ref{app:external}).
\begin{itemize}
\item[E1] Pricing and ordering ($d=4$): two products with prices and order quantities; gamma demands whose means depend on both prices and whose shapes depend on the product's own price, linked by a Gaussian copula; newsvendor profit with salvage and goodwill cost.
\item[E2] Portfolio with market impact and crowding ($d=5$, budget constraint): position size lowers expected returns and raises volatility; concentration raises the probability of a crash regime with lower returns and higher volatility.
\item[E3] Service system with induced demand ($d=3$): faster service attracts more customers; waiting times have an atom at zero and an exponential tail.
\item[E4] Project crashing ($d=5$): crashing shortens log-normal activity durations but makes them less predictable; the loss is the convex crashing cost plus a penalty for exceeding the deadline.
\item[E5] Insurance deductibles ($d=4$): higher deductibles reduce claim frequency and paid severity but also premium income; a catastrophe regime multiplies all losses.
\item[E6] Preventive maintenance ($d=3$): maintenance postpones Weibull failure times and makes them more regular; the loss is maintenance cost plus downtime cost.
\end{itemize}
E1--E3 were used during the development of the method and are therefore reported as a development set; E4--E6 serve as held-out problems (Section~\ref{sec:preregistration}).

\section{Experimental protocol}\label{sec:protocol}

\subsection{Budget accounting}\label{sec:budget}

Every simulator call made by a method, whether for design, surrogate training, candidate evaluation, reranking, pilot sampling, local search or selection, is debited from a single counter capped at $B$. When the cap would be exceeded, the method stops and returns its incumbent. Final evaluations and reference computations are not counted. All methods use essentially the whole budget: the median share used is between 96.5\% and 99.9\% for every method and family.

\subsection{Competitors and shared components}\label{sec:competitors}

All methods share the space-filling design sampler of Section~\ref{sec:design} and the local refinement of Section~\ref{sec:refine}, to which each method passes its own ordered list of starting points. Differences between methods therefore come from how they spend the global phase and which starts they produce (Table~\ref{tab:methods}). The refinement runs only if at least $2n_s+n_\ell(d+1)$ draws remain after its pilot sample. This condition is never met for GP-BO at $B=2000$ and for SGD-CVaR at $B\le 5000$, and it fails in 18\% of the CEM runs at $B=2000$; these methods then return the best point of their own phase. Settings that scale with the budget are listed in Appendix~\ref{app:settings}.

\begin{table}[t]
\caption{Methods compared. All methods use the same design sampler and, whenever enough budget remains, the same local refinement (Section~\ref{sec:refine}), which receives the listed starting points and uses the remaining budget}
\label{tab:methods}
\begin{tabular*}{\textwidth}{@{\extracolsep\fill}lp{0.38\textwidth}cp{0.24\textwidth}@{}}
\toprule
Method & Global phase & Share of $B$ & Starts passed to refinement \\
\midrule
ACRS & residual model, surrogate multistart, focused augmentation, reranking & 15--31\% & up to 8 surrogate optima in oracle order \\
GP-BO & Gaussian process on noisy estimates of $J$, expected improvement & 75\% & 3 best posterior means, best observed \\
CEM & cross-entropy search with Gaussian sampling distribution & 33--61\% & best sampled point, final mean \\
KDE-SO & kernel-weighted conditional surrogate, differential evolution, polish & 20\% ($\le$2,600) & up to 5 polished surrogate optima \\
SGD-CVaR & Adam with common-random-number central differences & $\approx$85\% & best checkpoint overall and of each chain \\
Refine-only & none & 0\% & 3 random feasible points \\
\botrule
\end{tabular*}
\end{table}

GP-BO \cite{Jones1998,Huang2006,Frazier2018} evaluates 30 design points and then the maximiser of expected improvement under a Gaussian process with squared-exponential kernel with automatic relevance determination and a noise (nugget) term \cite[Sect.~5.1]{Rasmussen2006}; the posterior is updated after every evaluation, the hyperparameters are re-estimated every ten evaluations by maximising the marginal likelihood, and expected improvement is maximised by L-BFGS-B from the two best of 1,000 random candidates. Each evaluation uses $\max(50,\lfloor 1000B/260000\rceil)$ fresh draws, and the loop stops when 75\% of the budget is spent; at $B=2000$ this leaves only the initial design. The recommendation is the evaluated point with the best posterior mean. CEM \cite{Rubinstein2004,DeBoer2005} runs ten iterations of a Gaussian cross-entropy search (elite fraction 0.15, smoothing 0.6, minimum standard deviation 0.015) after five warm-up evaluations. KDE-SO spends $\min(2600,0.2B)$ single draws on a design, estimates the conditional distribution by Gaussian kernel weights in decision space with a normal-reference bandwidth of the form $\mathrm{sd}_j\,N^{-1/(d+4)}$ \cite{Silverman1986}, evaluates $J$ on 600 resampled and jittered scenarios, minimises this surrogate by differential evolution \cite{Storn1997,Mullen2011} (population $\max(85,10d)$, 55 generations, current-to-$p$-best strategy \cite{Zhang2009}, $F=0.65$, crossover rate $0.85$) and polishes the best member and the 29 best members of the final population by L-BFGS-B. SGD-CVaR runs Adam \cite{Kingma2015} (moment parameters 0.9 and 0.999) with step size $0.05(1+0.001t)^{-1/2}$ on central differences with step 0.005 computed from 20 common draws, with two warm-up chains followed by a main chain from the better one, and spends about 85\% of the budget in this way plus checkpoint estimates on 60 draws; it passes the best checkpoint overall and the best checkpoint of each chain to the refinement. Refine-only has no global phase; it serves as a control for the global search phase.

\subsection{Performance measure and reference optima}\label{sec:regret}

For a returned decision $\hat x$, the normalised regret is
\begin{equation}
  \mathrm{regret}(\hat x)=100\,\frac{\tilde J(\hat x)-\tilde J(x^{\mathrm{ref}})}{\bar J-J^{\mathrm{ref}}},
  \label{eq:regret}
\end{equation}
where $x^{\mathrm{ref}}$ is a reference optimum with high-precision objective value $J^{\mathrm{ref}}$, $\bar J$ is the mean objective of uniformly random feasible decisions, and $\tilde J$ is the objective evaluated on the common evaluation sample of the replication (the exact objective in the analytic family). Thus 0 corresponds to the reference and 100 to the mean objective of random feasible decisions; this is the usual normalisation of solution accuracy in benchmarking of optimisation algorithms, with the best known value in place of the unknown optimum \cite{Beiranvand2017}. In the analytic family, $x^{\mathrm{ref}}$ is the best end point of 60 L-BFGS-B runs on the exact objective started from the 60 best of 4,000 Latin hypercube points, and $\bar J$ is the mean over these 4,000 points. For E1--E6, L-BFGS-B is run on a sample-path objective with 60,000 antithetic draws from the 30 best of 400 space-filling points, and the best of the five best end points on 200,000 independent draws is retained. For DGP1 and DGP2 the procedure uses 20,000 antithetic draws, 60 starts selected from about 1,300 feasible, boundary and face points, and verification of the ten best end points on 400,000 draws. In both cases $\bar J$ is estimated from 300 random feasible decisions with 20,000 common draws. Within a replication, all solutions and $x^{\mathrm{ref}}$ are evaluated on the same 20,000 (DGP) or 100,000 (E1--E6) draws, so the comparisons are paired, and the regret can be slightly negative when a solution beats the reference on the evaluation sample.

\subsection{Development and held-out evaluation}\label{sec:preregistration}

The method was developed on DGP1 and DGP2, on analytic instances and on E1--E3; these benchmarks are therefore not independent test sets, and E1--E3 are reported separately as a development set. The configuration of ACRS and of the shared refinement was fixed before the final experiments, and the refinement rule was chosen on separate development instances by the mean regret of several methods, not of ACRS alone. The 60 analytic test instances were generated anew, and the held-out problems E4--E6 were specified before any method was run on them, with parameters set using only reference optima.

\subsection{Statistical analysis}\label{sec:stats}

For each family, the unit of analysis is an instance (analytic family) or a problem--budget or DGP--$\lambda$--budget setting (other families), summarised by the mean regret over its replications. We report, for ACRS against each competitor, the numbers of units on which ACRS is better, tied (difference within 0.5 regret points) and worse, and a two-sided paired Wilcoxon signed-rank test on the unit means \cite{Wilcoxon1945,Hollander2014}, Holm-adjusted over the five competitors within each family \cite{Holm1979}. We also compare ACRS with the best competitor selected after the fact for each unit. This comparison is exploratory: the selection favours the competitors, and its $p$-value is not adjusted. Within a setting, replications are paired through common random numbers in the evaluation, and pairwise tests across replications are likewise Holm-adjusted over the five comparisons. No adjustment is made across families or settings. Intervals in the figures are 95\% percentile bootstrap intervals with 2,000 resamples \cite{Efron1993}. Normalised regret depends on $\bar J$, so values are comparable within a unit but only roughly across families.

\subsection{Software and hardware}\label{sec:software}

All methods were implemented in R 4.5.1 \cite{RCore2025} with the packages data.table 1.17.8, DEoptim 2.2.8 \cite{Mullen2011} and ggplot2 4.0.3; the forest-based comparators of Section~\ref{sec:ablation} use grf 2.5.0 and drf 1.3.1. All figures were produced in R with ggplot2 and grid. Each run used one core of an Apple M4 Pro (24\,GB RAM, macOS 15.1), with replications in parallel. The code, the raw results and the scripts that generate every table and figure are available from the author on reasonable request.

\section{Results}\label{sec:results}

\subsection{Overview}\label{sec:overview}

Table~\ref{tab:h2h} compares ACRS with each competitor in each family. ACRS has the lowest mean regret in the analytic family (11.8), on E1--E3 (10.4) and on DGP1/DGP2 (0.4). On the held-out problems E4--E6 its mean regret (13.5) equals that of Refine-only to one decimal and is lower than that of KDE-SO (17.1); neither difference is significant. After Holm adjustment, ACRS is significantly better than every competitor in the analytic family and on E1--E3, than GP-BO, SGD-CVaR and Refine-only on DGP1/DGP2, and than SGD-CVaR on E4--E6; on E4--E6 its advantage over GP-BO and CEM is borderline (adjusted $p=0.051$), and on DGP1/DGP2 the differences to CEM and KDE-SO are not significant (adjusted $p=0.081$). The best competitor selected after the fact for each unit is not significantly better than ACRS in any family; in the analytic family ACRS is better on 26 instances and worse on 13 ($p=0.059$), and on E1--E3 this oracle selection is significantly worse than ACRS.

\begin{table}[t]
\caption{Head-to-head comparison of ACRS with each competitor, by family. Units are instances (analytic family) or settings (other families), summarised by their mean normalised regret. Better/tie/worse count units on which ACRS has lower, equal (within 0.5) or higher mean regret; $p$ is a two-sided paired Wilcoxon signed-rank test on unit means, Holm-adjusted over the five competitors within each family (unadjusted for the best competitor per unit)}
\label{tab:h2h}
\begin{tabular*}{\textwidth}{@{\extracolsep\fill}lrrrrr@{}}
\toprule
Opponent & Mean regret & \multicolumn{3}{c}{ACRS better / tie / worse} & $p$ \\
\midrule
\multicolumn{6}{@{}l}{\emph{Analytic family, 60 instances; ACRS mean regret 11.8}}\\
GP-BO & 29.4 & 48 & 7 & 5 & $<$0.001 \\
CEM & 22.3 & 40 & 14 & 6 & $<$0.001 \\
KDE-SO & 26.1 & 38 & 17 & 5 & $<$0.001 \\
SGD-CVaR & 42.3 & 56 & 2 & 2 & $<$0.001 \\
Refine-only & 22.9 & 39 & 12 & 9 & $<$0.001 \\
Best competitor per unit$^{a}$ & 12.6 & 26 & 21 & 13 & 0.059 \\
\midrule
\multicolumn{6}{@{}l}{\emph{E1--E3 (development set), 9 settings; ACRS mean regret 10.4}}\\
GP-BO & 28.9 & 9 & 0 & 0 & 0.046 \\
CEM & 31.3 & 9 & 0 & 0 & 0.046 \\
KDE-SO & 29.6 & 8 & 1 & 0 & 0.046 \\
SGD-CVaR & 39.3 & 9 & 0 & 0 & 0.046 \\
Refine-only & 22.2 & 8 & 1 & 0 & 0.046 \\
Best competitor per unit$^{a}$ & 17.3 & 7 & 2 & 0 & 0.013 \\
\midrule
\multicolumn{6}{@{}l}{\emph{E4--E6 (held-out set), 9 settings; ACRS mean regret 13.5}}\\
GP-BO & 35.7 & 8 & 0 & 1 & 0.051 \\
CEM & 26.2 & 7 & 2 & 0 & 0.051 \\
KDE-SO & 17.1 & 6 & 0 & 3 & 0.814 \\
SGD-CVaR & 61.3 & 9 & 0 & 0 & 0.046 \\
Refine-only & 13.5 & 5 & 0 & 4 & 0.814 \\
Best competitor per unit$^{a}$ & 11.5 & 4 & 1 & 4 & 1.000 \\
\midrule
\multicolumn{6}{@{}l}{\emph{DGP1 and DGP2, 18 settings; ACRS mean regret 0.4}}\\
GP-BO & 2.2 & 10 & 8 & 0 & 0.008 \\
CEM & 1.7 & 3 & 15 & 0 & 0.081 \\
KDE-SO & 1.0 & 7 & 11 & 0 & 0.081 \\
SGD-CVaR & 24.6 & 15 & 3 & 0 & 0.002 \\
Refine-only & 1.4 & 7 & 11 & 0 & 0.014 \\
Best competitor per unit$^{a}$ & 0.7 & 3 & 15 & 0 & 0.760 \\
\botrule
\end{tabular*}
\footnotetext[a]{For each unit, the competitor with the lowest mean regret on that unit, chosen after the fact.}
\end{table}

\subsection{Analytic family}\label{sec:res-analytic}

\begin{figure}[t]
\centering
\includegraphics{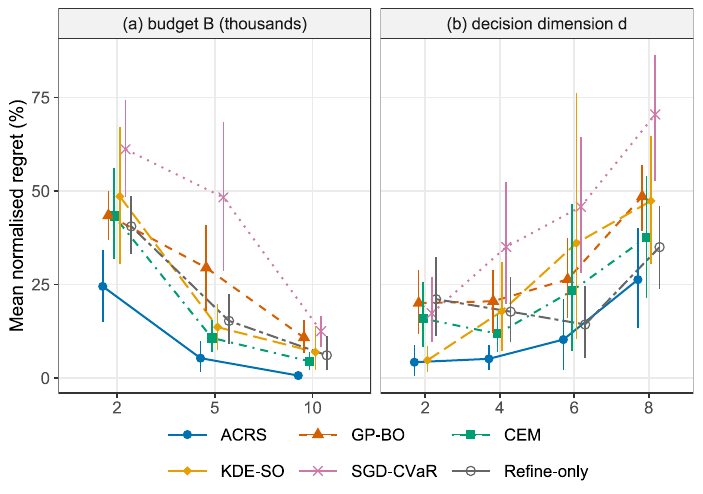}
\caption{Analytic family: mean normalised regret by (a) budget (thousands of simulator draws) and (b) decision dimension, with 95\% bootstrap intervals over instances (60 instances, 4 replications each). Methods are distinguished by marker shape and line type}
\label{fig:analytic}
\end{figure}

Fig.~\ref{fig:analytic} and Table~\ref{tab:breakdown} break the analytic results down by budget, dimension and jump regime. ACRS has the lowest mean regret in every row of Table~\ref{tab:breakdown}. Its lead is largest at the smallest budget (24.5 against at least 40.6 at $B=2000$) and persists at the largest (0.7 against at least 4.5 at $B=10{,}000$), although the absolute differences shrink with the budget. The difference between the two conditional-distribution methods tends to grow with the dimension: in two dimensions KDE-SO is almost as good as ACRS (4.7 against 4.2), because a kernel estimate has enough neighbours, while at $d=6$ and $d=8$ its mean regret (36.1 and 47.3) is 3.5 and 1.8 times that of ACRS. A regression of the instance-level difference between KDE-SO and ACRS on the instance factors confirms this: the difference grows by 3.5 regret points per unit of dimension ($p=0.017$) and by 6.4 points per unit of the strength $\delta$ of decision dependence ($p=0.038$), and it falls by 10.7 points per doubling of the budget ($p=0.002$), whereas the number of wells ($p=0.73$) and the jump regime ($p=0.36$) have no significant effect. Against Refine-only, the value of the global phase is largest in two dimensions (16.9 points) and smaller at $d=6$ and $d=8$ (4.0 and 8.7 points), where 15--20\% of a small budget can resolve the landscape only coarsely. The jump regime, which makes sample paths discontinuous, changes the regret of ACRS only slightly (13.5 against 11.4), and ACRS remains the best method on both groups.

\begin{table}[t]
\caption{Analytic family: mean normalised regret by budget, dimension and jump regime ($n$ = number of instances). The lowest value in each row is in bold}
\label{tab:breakdown}
\begin{tabular*}{\textwidth}{@{\extracolsep\fill}lrrrrrrr@{}}
\toprule
 & $n$ & ACRS & GP-BO & CEM & KDE-SO & SGD-CVaR & Refine-only \\
\midrule
$B=2000$ & 25 & \textbf{24.5} & 43.5 & 43.3 & 48.6 & 61.2 & 40.6 \\
$B=5000$ & 16 & \textbf{5.3} & 29.5 & 10.7 & 13.6 & 48.3 & 15.3 \\
$B=10000$ & 19 & \textbf{0.7} & 10.8 & 4.5 & 6.9 & 12.5 & 6.1 \\
\midrule
$d=2$ & 16 & \textbf{4.2} & 20.0 & 15.9 & 4.7 & 17.3 & 21.1 \\
$d=4$ & 16 & \textbf{5.1} & 20.5 & 11.8 & 17.9 & 35.1 & 17.7 \\
$d=6$ & 11 & \textbf{10.3} & 26.4 & 23.3 & 36.1 & 45.8 & 14.3 \\
$d=8$ & 17 & \textbf{26.3} & 48.6 & 37.6 & 47.3 & 70.5 & 35.0 \\
\midrule
No jump & 48 & \textbf{11.4} & 29.2 & 20.5 & 27.0 & 40.2 & 23.2 \\
Jump & 12 & \textbf{13.5} & 30.3 & 29.7 & 22.4 & 50.8 & 21.8 \\
\botrule
\end{tabular*}
\end{table}

\subsection{External problems}\label{sec:res-external}

\begin{figure}[t]
\centering
\includegraphics{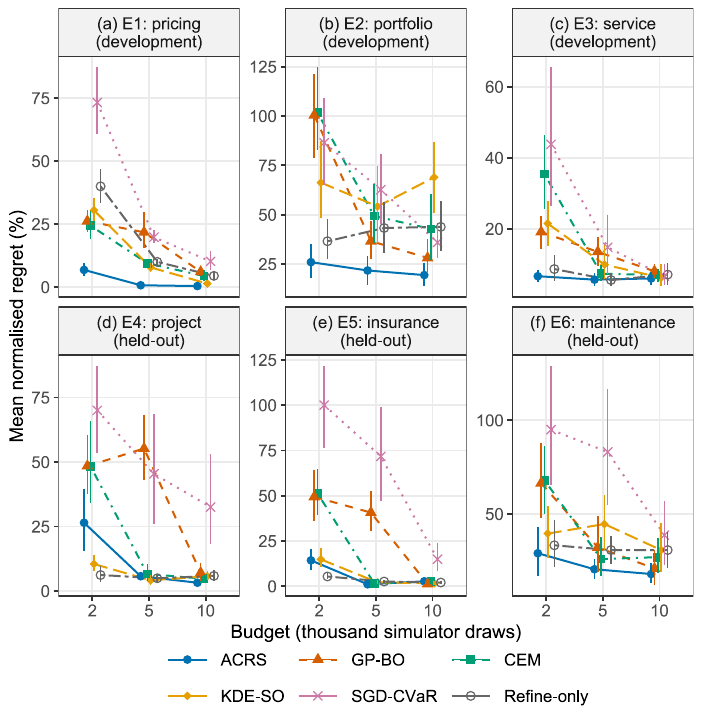}
\caption{External problems: mean normalised regret by budget (thousands of simulator draws) with 95\% bootstrap intervals over 20 replications; (a)--(c) development set E1--E3, (d)--(f) held-out set E4--E6. Methods are distinguished by marker shape and line type. The vertical scales differ between panels}
\label{fig:external}
\end{figure}

Fig.~\ref{fig:external} and Table~\ref{tab:external} report the external problems. On the development set E1--E3, ACRS has the lowest median regret in seven of nine settings and is significantly better than every competitor on E1 at all three budgets. E2, the portfolio problem with a decision-dependent crash regime, is the hardest problem for all methods; ACRS reaches a median regret of 16--20 against 23--104 for the competitors. On E3 at $B\ge 5000$, ACRS is statistically indistinguishable from all competitors except GP-BO at $B=5000$.

The held-out set is less favourable to ACRS. On E4 and E5 at $B=2000$, Refine-only is significantly better than ACRS (median regret 4.6 and 4.2 against 13.0 and 10.6). On these problems, spending the entire small budget on local descent is more effective: random starts already lead to good basins, and the 21\% of the budget (median) that ACRS spends on its global phase at this budget is then missing from the refinement. At $B=5000$ and $10{,}000$, ACRS has the lowest median in two of the four settings on E4 and E5 and is never significantly worse than a competitor. E6 is very flat: the mean objective of random decisions exceeds the optimum by only 2\%, so a regret of 20 corresponds to a relative gap of about 0.4\%. Here ACRS has the lowest median at $B=2000$ and $5000$ and the lowest median relative gap over all budgets (0.31\% against 0.53--1.08\%), although few of the differences are significant.

\begin{table}[t]
\caption{External problems: median normalised regret over 20 replications. Bold: lowest median in the row. $^{+}$/$^{-}$: the competitor is significantly worse/better than ACRS (paired Wilcoxon test over replications, Holm-adjusted over five comparisons, level 0.05)}
\label{tab:external}
\begin{tabular*}{\textwidth}{@{\extracolsep\fill}llrrrrrr@{}}
\toprule
Problem & $B$ (k) & ACRS & GP-BO & CEM & KDE-SO & SGD-CVaR & Refine-only \\
\midrule
\multicolumn{8}{@{}l}{\emph{Development set}}\\
E1 & 2 & \textbf{3.8} & 26.7$^{+}$ & 21.3$^{+}$ & 29.3$^{+}$ & 74.4$^{+}$ & 39.3$^{+}$ \\
E1 & 5 & \textbf{0.4} & 16.2$^{+}$ & 10.1$^{+}$ & 9.6$^{+}$ & 18.0$^{+}$ & 10.1$^{+}$ \\
E1 & 10 & \textbf{0.3} & 5.0$^{+}$ & 2.5$^{+}$ & 0.5$^{+}$ & 9.8$^{+}$ & 3.4$^{+}$ \\
E2 & 2 & \textbf{20.1} & 103.5$^{+}$ & 100.3$^{+}$ & 59.5$^{+}$ & 68.5$^{+}$ & 35.7 \\
E2 & 5 & \textbf{16.8} & 31.3$^{+}$ & 38.5$^{+}$ & 36.3$^{+}$ & 64.5$^{+}$ & 39.1$^{+}$ \\
E2 & 10 & \textbf{15.9} & 23.4 & 24.4 & 76.8$^{+}$ & 34.0 & 40.3$^{+}$ \\
E3 & 2 & 7.3 & 22.1$^{+}$ & 30.0$^{+}$ & 19.3$^{+}$ & 28.3$^{+}$ & \textbf{6.8} \\
E3 & 5 & \textbf{4.6} & 12.7$^{+}$ & 6.8 & 5.9 & 7.5 & 5.1 \\
E3 & 10 & 5.1 & 6.5 & 6.5 & \textbf{4.1} & 4.7 & 4.6 \\
\midrule
\multicolumn{8}{@{}l}{\emph{Held-out set}}\\
E4 & 2 & 13.0 & 41.5 & 41.9 & 9.9 & 60.0$^{+}$ & \textbf{4.6}$^{-}$ \\
E4 & 5 & 5.5 & 46.8$^{+}$ & 3.8 & \textbf{2.5} & 20.0$^{+}$ & 3.9 \\
E4 & 10 & \textbf{3.5} & 3.6 & 3.8 & 4.7 & 16.4$^{+}$ & 4.2 \\
E5 & 2 & 10.6 & 38.9$^{+}$ & 37.6$^{+}$ & 11.4 & 109.6$^{+}$ & \textbf{4.2}$^{-}$ \\
E5 & 5 & \textbf{0.7} & 31.4$^{+}$ & 1.0 & 1.3 & 66.5$^{+}$ & 2.8$^{+}$ \\
E5 & 10 & 1.3 & 0.9 & 1.7 & \textbf{0.7} & 14.4 & 1.1 \\
E6 & 2 & \textbf{16.2} & 59.9$^{+}$ & 64.8$^{+}$ & 29.3 & 64.9$^{+}$ & 28.1 \\
E6 & 5 & \textbf{16.5} & 23.5 & 17.8 & 41.2 & 64.3$^{+}$ & 30.8 \\
E6 & 10 & 13.7 & \textbf{11.3} & 21.8 & 19.1 & 20.0 & 27.1 \\
\botrule
\end{tabular*}
\end{table}

\subsection{DGP1 and DGP2}\label{sec:res-dgp}

On DGP1 and DGP2, all methods except SGD-CVaR come close to the reference (Table~\ref{tab:dgp} in Appendix~\ref{app:tables}). ACRS has the lowest mean regret (0.4) and is significantly better than GP-BO, SGD-CVaR and Refine-only, whereas the differences to CEM (1.7) and KDE-SO (1.0) are not significant after adjustment. The differences arise mainly at $B=5000$, where the median relative gap to the reference is 0.5\% for ACRS, 2.1\% for KDE-SO, 2.3\% for CEM, 7.9\% for GP-BO and 8.3\% for Refine-only. At $B=30{,}000$ several methods are slightly better than ACRS on DGP1, significantly so only KDE-SO at $\lambda=0.9$. Negative regret occurs in 7--25\% of the runs of each method, so many of the remaining differences are at the level of the evaluation noise. Apart from SGD-CVaR and, at $B=5000$, GP-BO and Refine-only, these benchmarks separate the methods little.

\subsection{Phase diagnostics}\label{sec:phases}

\begin{figure}[t]
\centering
\includegraphics[width=0.8\linewidth]{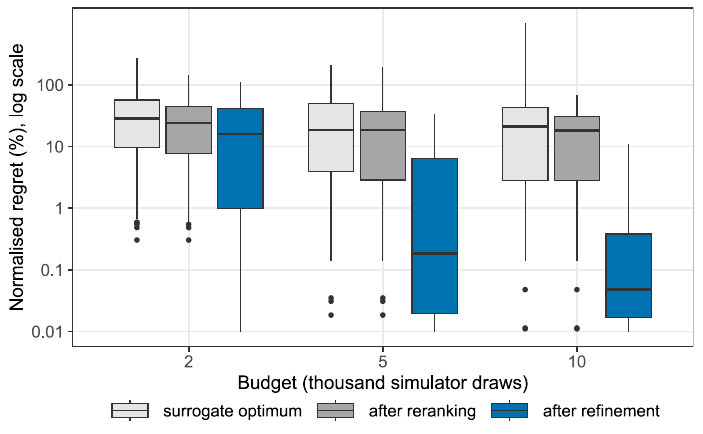}
\caption{Analytic family: exact normalised regret of ACRS after the surrogate search, after oracle reranking and after the local refinement, by budget (240 runs; boxes show quartiles, whiskers 1.5 interquartile ranges; values below 0.01 are shown at 0.01)}
\label{fig:phases}
\end{figure}

Because the analytic family has an exact objective, the regret of ACRS can be measured after each phase (Fig.~\ref{fig:phases}). The surrogate optimum alone has a high regret: its median is 23.9 over all instances, consistent with the misspecification discussed in Remark~\ref{rem:consistency}, since the tail index and the correlation in this family depend on the decision. Oracle reranking lowers the median to 21.0, and the refinement to 0.5; at $B=10{,}000$ the medians are 21.0, 18.4 and below 0.1. The global phase thus contributes mainly a good starting point: from three random starts, the same refinement reaches a median regret of 14.3 (Table~\ref{tab:ablation}). At these budgets the refinement typically completes at most one local run (the median number of local runs is one for every method except SGD-CVaR, for which it is zero), so the first start largely determines the result, which explains the effect of reranking. The surrogate usually has few distinct candidates (median one or two).

\subsection{Ablation and choice of the conditional model}\label{sec:ablation}

\begin{figure}[t]
\centering
\includegraphics[width=0.8\linewidth]{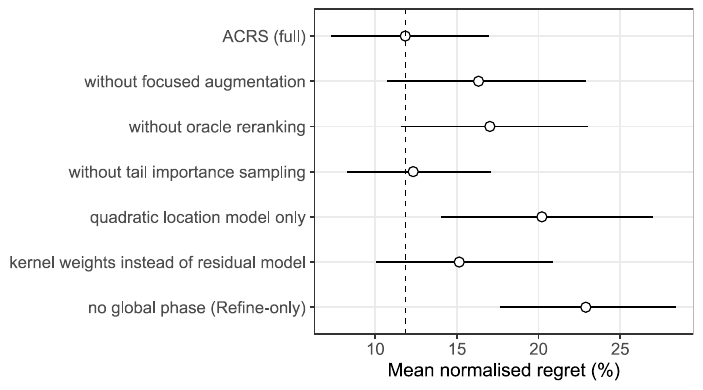}
\caption{Ablation of ACRS on the 60 analytic instances: mean normalised regret with 95\% bootstrap intervals over instances. The dashed line marks the full method}
\label{fig:ablation}
\end{figure}

Table~\ref{tab:ablation} and Fig.~\ref{fig:ablation} report variants of ACRS on the 60 analytic instances; the variants share the random-number streams of the main runs. Removing focused augmentation raises the mean regret from 11.8 to 16.3 ($p=0.005$) and removing reranking to 17.0 ($p=0.004$). Replacing the flexible location model by a quadratic one gives 20.2 ($p<0.001$), and replacing the residual model by kernel weights within the same pipeline gives 15.1 ($p=0.11$). Removing tail importance sampling from the refinement changes little (12.3, $p=0.33$); we retain it in the shared refinement because the defensive mixture bounds the importance weights (Proposition~\ref{prop:is}(a)). Removing the whole global phase (Refine-only) raises the mean regret to 22.9 ($p<0.001$) and the median from 0.5 to 14.3.

\begin{table}[t]
\caption{Ablation of ACRS on the 60 analytic instances (4 replications each). $p$: two-sided paired Wilcoxon test against the full method on instance means, Holm-adjusted over the six variants}
\label{tab:ablation}
\begin{tabular*}{\textwidth}{@{\extracolsep\fill}lrrr@{}}
\toprule
Variant & Mean regret & Median regret & $p$ \\
\midrule
ACRS (full) & 11.8 & 0.5 & -- \\
without focused augmentation & 16.3 & 1.0 & 0.005 \\
without oracle reranking & 17.0 & 2.2 & 0.004 \\
without tail importance sampling & 12.3 & 0.7 & 0.333 \\
quadratic location model only & 20.2 & 4.5 & $<$0.001 \\
kernel weights instead of residual model & 15.1 & 1.0 & 0.112 \\
no global phase (Refine-only) & 22.9 & 14.3 & $<$0.001 \\
\botrule
\end{tabular*}
\end{table}

A preliminary experiment on DGP1 and DGP2 ($\lambda=0.7$) compared conditional models by the quality of the surrogate minimiser when no further simulator draws are spent (Table~\ref{tab:surrogate}). For each of 15 replications, $N$ single draws were taken on the feasible design, each model was fitted, its surrogate was minimised by differential evolution and L-BFGS-B, and the minimiser was evaluated on 20,000 draws against the reference. The residual model with a quadratic location gave the smallest relative gap in all four cases and was the fastest. Generalised \cite{Athey2019} and distributional \cite{Cevid2022} random forest weights gave gaps 1.5 to 5.5 times larger and took 20 to 50 times longer; combining residual scenarios with forest weights did not help. This experiment motivated the choice of the residual model; the radial-basis features of the location model reduce the regret further (Table~\ref{tab:ablation}).

\begin{table}[t]
\caption{Choice of conditional model (preliminary experiment): median relative gap (\%) between the objective at the surrogate minimiser and at the reference optimum, without any simulator draws after the $N$ design draws (15 replications per cell), and median fitting and search time}
\label{tab:surrogate}
\begin{tabular*}{\textwidth}{@{\extracolsep\fill}lrrrrr@{}}
\toprule
 & \multicolumn{2}{c}{DGP1} & \multicolumn{2}{c}{DGP2} & \\
\cmidrule(lr){2-3}\cmidrule(lr){4-5}
Conditional model & $N=1000$ & $N=5000$ & $N=1000$ & $N=5000$ & Time (s) \\
\midrule
Residual, quadratic location & \textbf{18.5} & \textbf{6.1} & \textbf{44.6} & \textbf{27.0} & 1.7 \\
Residual with forest weights & 20.7 & 14.8 & \textbf{44.6} & 35.1 & 78.0 \\
Kernel weights & 31.5 & 38.2 & 74.8 & 54.2 & 2.2 \\
Generalised random forest & 48.1 & 29.4 & 72.4 & 63.6 & 85.0 \\
Distributional random forest & 52.2 & 33.6 & 69.9 & 66.0 & 41.3 \\
\botrule
\end{tabular*}
\end{table}

\subsection{Computational cost}\label{sec:cost}

The simulators used here are cheap, so wall-clock times reflect the overhead of each method. The median time per run was 1.4\,s for ACRS, 1.5\,s for GP-BO, 3.8\,s for KDE-SO and below 0.05\,s for CEM, SGD-CVaR and Refine-only in the analytic family, and 8.6\,s, 26.4\,s and 8.2\,s for ACRS, GP-BO and KDE-SO on DGP1/DGP2. When simulator draws are expensive, their cost is likely to dominate the overhead of each method, and it is similar across methods because all of them use nearly the full budget.

\section{Discussion and limitations}\label{sec:discussion}

ACRS is most useful when the budget is small relative to the dimension and the landscape has several basins, and when decision dependence acts mainly on the location and scale of the uncertain parameters, so that the residual model captures the large-scale structure of $J$. In two dimensions, a kernel estimate is almost as good. When random starts already reach good basins, as on E4 and E5 at the smallest budget, a pure local method is better because the global phase uses budget without finding a better basin. On very flat objectives such as E6, normalised regret magnifies small relative differences, and few of them are significant at these budgets.

Several limitations qualify these conclusions. First, the benchmarks are synthetic or stylised; they were chosen to vary the structure of decision dependence, but they are not industrial simulators. Second, the decision dimension is at most eight and the budgets range from 2,000 to 30,000 draws; we have not studied high-dimensional problems, where the quadratic features would have to be replaced by additive or sparse ones. Third, Assumption~\ref{ass:sim} requires control over the simulator input; common random numbers need only a seed, but importance sampling in the input space needs the input distribution. The ablation shows that ACRS does not depend on importance sampling, but the shared refinement of all methods does use it. Fourth, the location--scale model cannot represent decision-dependent tail shapes or regime switches. ACRS uses the surrogate only to propose candidates, and the phase diagnostics show that this can be effective on the analytic family; a heteroscedastic model of the residual shape, for example by quantile regression of the standardised residuals, could reduce the bias. Fifth, the refinement typically completes one local run at these budgets, so budget allocation between candidates, for example by optimal computing budget allocation, could help all methods. Sixth, E1--E3 influenced the development, and normalised regret depends on the mean objective of random decisions; we therefore report a held-out set, report relative gaps where the normalisation is uninformative, and make all raw results available on request.

\section{Conclusions}\label{sec:conclusions}

We introduced adaptive conditional residual sampling (ACRS) for minimising the sum of expected cost and weighted CVaR under decision-dependent uncertainty with a fixed simulation budget. ACRS learns the decision-dependent distribution from single draws at a space-filling design by pooling standardised residuals, searches the resulting deterministic surrogate for several basins, and uses simulator evaluations for reranking with common random numbers and for a sample-path refinement with tail importance sampling. The theory bounds the optimality gap of the surrogate by the Wasserstein error of its scenario distribution and establishes consistency of the refinement estimator. Under strictly equal budget accounting, with a shared design sampler and a local refinement routine available to every method whenever enough budget remained, ACRS had the lowest mean regret on 60 new analytic instances, on two six-dimensional benchmarks and on three development problems, and its mean regret was comparable to that of pure local refinement on three held-out problems. No competitor, including the best one chosen after the fact for each unit, was significantly better on average in any family, but pure local refinement was better on two held-out problems at the smallest budget. Future work includes adaptive allocation of the budget between the global and local phases, residual models with decision-dependent shape, and extensions to higher dimensions and constrained problems.

\backmatter

\section*{Statements and Declarations}

\textbf{Funding.} The author did not receive support from any organisation for the submitted work.

\noindent\textbf{Competing interests.} The author has no relevant financial or non-financial interests to disclose.

\noindent\textbf{Ethics approval and consent.} Not applicable.

\noindent\textbf{Data and code availability.} All data analysed in this study were generated by the simulation models specified in full in Section~\ref{sec:benchmarks} and Appendices~\ref{app:analytic}--\ref{app:external}. The R code of all methods and benchmark problems, the raw per-run results of all experiments, the reference optima and the scripts that reproduce every table and figure are available from the corresponding author on reasonable request.

\noindent\textbf{Use of generative AI.} The author used Claude Opus 5 (Anthropic) for assistance with manuscript editing and code development. All outputs were reviewed and verified by the author, who takes full responsibility for the content of the article.

\noindent\textbf{Author contributions.} M.T. Kurbucz is the sole author and was responsible for the conception, methodology, software, analysis and writing of this work.

\bibliography{ACRS}

\begin{appendices}

\section{Analytic family}\label{app:analytic}\setcounter{table}{0}

Each instance is generated from $(d,K,\delta)$ and a random seed. The well width is $r=0.07^{1/d}/1.86$, so that the volume share of the wells is similar in every dimension. The first well centre is uniform on $[0.15,0.85]^d$; further centres are drawn from the same box and accepted if their distance to all accepted centres is at least $2.5r$. The depths are $h_1=1$ and $h_k\sim U(0.7,1)$ for $k\ge 2$; the loadings $A\in\R^{3\times K}$ have entries $U(0.5,1.5)$; the extra log-scales are $e_k\sim U(-0.3,1.2)$; $G\in\R^{3\times d}$ has entries $\mathcal{N}(0,1/d)$; $u,v\sim\mathcal{N}(0,I_d)$; and $x_c\sim U(0.2,0.8)^d$. The extra log-scale enters every component of $\sigma(x)$, so some wells carry more risk than others and the deepest well of the mean is not always the best basin of $J$.

The simulator computes $Y=Z_{1:3}/\sqrt{\chi/\nu(x)}$ from the input $Z\in\R^4$, with $\chi=F^{-1}_{\chi^2_{\nu(x)}}(\Phi(Z_4))$, and returns $W=\mu(x)+\sigma(x)\odot(\Lambda(x)Y)$, where $\Lambda(x)$ is the Cholesky factor of the equicorrelation matrix with parameter $\varrho(x)$.

Since $\mathbf{1}^\top(\sigma\odot\Lambda Y)=(\Lambda^\top\sigma)^\top Y$ and $Y$ is spherical, this sum equals $s_L(x)T_{\nu(x)}$ in distribution with $s_L(x)^2=\sigma^\top\Lambda\Lambda^\top\sigma=(1-\varrho)\|\sigma\|^2+\varrho(\mathbf{1}^\top\sigma)^2$, and \eqref{eq:analyticJ} follows from $\E T_\nu=0$ ($\nu\ge 2.5$) and the closed form of the CVaR of the Student-$t$ distribution.

In the jump variant, with a fifth input $Z_5$, the first component of $W$ is increased by $\Delta(x)=3s_L(x)$ if $\Phi(Z_5)<\pi(x)$, where $\pi(x)=0.02+0.06/(1+\exp\{-4\delta\,\tilde u^\top(x-\frac12\mathbf{1})/\sqrt d\})$ and $\tilde u\sim\mathcal{N}(0,I_d)$. The loss is $m_L(x)+s_L(x)T_\nu+\Delta(x)\xi$ with $\xi\sim\mathrm{Bernoulli}(\pi(x))$ independent of $T_\nu$, so $\E L=m_L+\pi\Delta$. With $F_\nu$ the distribution function of $t_\nu$ and $\varpi_\nu(t)=\E(T_\nu-t)_+=\frac{\nu+t^2}{\nu-1}f_\nu(t)-t\,(1-F_\nu(t))$, the value-at-risk $v$ solves $(1-\pi)F_\nu((v-m_L)/s_L)+\pi F_\nu((v-m_L-\Delta)/s_L)=\alpha$, and
\begin{equation*}
\CVaR_\alpha(L)=v+\frac{s_L}{1-\alpha}\Bigl[(1-\pi)\,\varpi_\nu\Bigl(\frac{v-m_L}{s_L}\Bigr)+\pi\,\varpi_\nu\Bigl(\frac{v-m_L-\Delta}{s_L}\Bigr)\Bigr].
\end{equation*}

\section{DGP1 and DGP2}\label{app:dgp}\setcounter{table}{0}

Throughout, $x_0=\max(0,1-\sum_{j=1}^5x_j)$ and $(z)_+=\max(0,z)$. Decisions are projected onto the feasible set by clipping to the bounds and, if $\sum_{j\le5}x_j>0.85$, rescaling $x_1,\ldots,x_5$ to sum to 0.85. The feasible design maps Latin hypercube uniforms $U_1,\ldots,U_5$ to the spacings of their order statistics, multiplied by 0.85 and randomly permuted, which is uniform on $\{x_{1:5}\ge 0,\sum x_j\le 0.85\}$; points with some $x_j>0.7$ are rejected, and $x_6$ is the sixth Latin hypercube coordinate.

In DGP1, a correlated normal vector $\tilde Z=\Lambda(x)Z$ with $\Lambda(x)\Lambda(x)^\top=R(x)$ is mapped to $U_j=\Phi(\tilde Z_j)$ and $W_j=\mu_j(x)+\sigma_j(x)F^{-1}_{t,\nu_j(x)}(U_j)$, with the parameters of Table~\ref{tbl:dgp1}. When $R(x)$ is not positive definite, its eigenvalues are floored at $10^{-6}$ without rescaling the resulting matrix to unit diagonal. This occurs only in DGP1, on approximately 0.05\% of the feasible set and away from the reference optima. The resulting diagonal entries exceed one by less than 0.03. With capacities $\kappa_1=0.45+0.80x_2+0.18x_0$ and $\kappa_2=0.40+0.75x_4+0.15x_0$ and the smoothed positive part $\mathrm{sp}(z)=\log(1+\mathrm{e}^{\min(8z,40)})/8$,
\begin{align*}
C(W,x)={}&2800\,W_2^2\bigl(1+0.65(W_1)_+\bigr)+1600\,W_4^2\bigl(1+0.45(W_5)_+\bigr)\\
&+10000\,\mathrm{sp}(W_3-\kappa_1)^2+6000\,\mathrm{sp}(W_5-\kappa_2)^2\\
&+2500(1-0.38x_3)(W_2)_+^{1.8}+1800(1-0.30x_5)(W_4)_+^{1.6}\\
&+14000\Bigl(\textstyle\sum_{j=1}^5x_j\Bigr)\max\bigl\{0.04,\;0.22-0.12x_6+0.06(W_1)_++0.04(W_5)_+\bigr\}\\
&+9500(x_1^2+0.85x_2^2+0.72x_3^2+0.68x_4^2+0.60x_5^2)\\
&+6500x_6^2+2800(x_1+x_2)x_6+2200(x_3+x_4)x_6 .
\end{align*}

\begin{table}[!htb]
\caption{Decision-dependent parameters of DGP1 (Gaussian copula with Student-$t$ marginals); $\sigma_j$ is the scale of the $t$ marginal}
\label{tbl:dgp1}
\begin{tabular*}{\textwidth}{@{\extracolsep\fill}ll@{}}
\toprule
Parameter & Expression \\
\midrule
$\mu_1(x)$ & $2.10 - 1.50\,x_6 - 1.10\,x_1 + 1.20\,x_0 + 0.75\,x_2 x_3 - 0.55\,x_1 x_6^2$ \\
$\mu_2(x)$ & $2.30 - 1.30\,x_1 - 1.70\,x_3 + 0.95\,x_0 - 0.65\,x_6^2 + 0.45\,x_2^2$ \\
$\mu_3(x)$ & $1.90 - 1.50\,x_2 - 1.05\,x_6 + 0.75\,x_0 + 0.55\,x_1 x_2 - 0.35\,x_3 x_6$ \\
$\mu_4(x)$ & $1.70 - 1.20\,x_4 - 0.90\,x_6 + 0.60\,x_0 + 0.40\,x_3 x_5 - 0.30\,x_2 x_4$ \\
$\mu_5(x)$ & $1.50 - 1.00\,x_5 - 0.80\,x_6 + 0.50\,x_0 + 0.35\,x_1 x_4 - 0.25\,x_3^2$ \\
\midrule
$\sigma_1(x)$ & $0.20 + 0.75(1-x_6)(1-x_1) + 0.30\,x_0$ \\
$\sigma_2(x)$ & $0.24 + 0.65(1-x_3) + 0.28(1-x_6) + 0.22\,x_0$ \\
$\sigma_3(x)$ & $0.22 + 0.60(1-x_2) + 0.20(1-x_6) + 0.16\,x_0$ \\
$\sigma_4(x)$ & $0.18 + 0.50(1-x_4) + 0.18(1-x_6) + 0.14\,x_0$ \\
$\sigma_5(x)$ & $0.16 + 0.45(1-x_5) + 0.15(1-x_6) + 0.12\,x_0$ \\
\midrule
$\nu(x)$ & $(3.0 + 2.5\,x_6,\; 3.0 + 2.0\,x_3,\; 3.0 + 1.5\,x_2,\; 3.5 + 2.0\,x_4,\; 3.5 + 1.5\,x_5)$ \\
\midrule
$R_{12}, R_{13}$ & $0.55 + 0.28(1-x_6)$, \quad $0.20 + 0.28(1-x_1)$ \\
$R_{14}, R_{15}$ & $0.15 + 0.20(1-x_2)$, \quad $0.10 + 0.18(1-x_3)$ \\
$R_{23}, R_{24}$ & $0.50 + 0.32(1-2x_2)$, \quad $0.20 + 0.22(1-x_4)$ \\
$R_{25}, R_{34}$ & $0.15 + 0.18(1-x_5)$, \quad $0.25 + 0.20(1-x_3)$ \\
$R_{35}, R_{45}$ & $0.18 + 0.16(1-x_4)$, \quad $0.30 + 0.25(1-2x_5)$ \\
\botrule
\end{tabular*}
\end{table}

DGP2 uses the same construction with $W_j=\exp\{\mu_j(x)+\sigma_j(x)\tilde Z_j\}$, where $\mu_j$ and $\sigma_j$ (floored at 0.05) are the location and scale of $\log W_j$ (Table~\ref{tbl:dgp2}). With holding capacities $\kappa^h_1=0.50+0.70x_1+0.20x_0$, $\kappa^h_2=0.45+0.65x_3+0.18x_0$ and shortage thresholds $\kappa^s_1=0.60+0.80x_2+0.25x_0$, $\kappa^s_2=0.55+0.75x_4+0.22x_0$,
\begin{align*}
C(W,x)={}&3500(W_1-\kappa^h_1)_+^2+2800(W_2-\kappa^h_2)_+^2\\
&+12000(\kappa^s_1-W_3)_+^{1.5}+9000(\kappa^s_2-W_5)_+^{1.5}\\
&+1800(1-0.40x_1)(W_1)_+^{1.6}+1400(1-0.35x_3)(W_4)_+^{1.4}\\
&+8000\Bigl(\textstyle\sum_{j=1}^5x_j\Bigr)\max\bigl\{0.05,\;0.18-0.10x_6+0.05(W_2)_+^2+0.03(W_5)_+^2\bigr\}\\
&+7500(x_1^2+0.80x_2^2+0.65x_3^2+0.60x_4^2+0.55x_5^2)+5000x_6^2\\
&+2200(x_1+x_2)x_6 .
\end{align*}

\begin{table}[!htb]
\caption{Decision-dependent parameters of DGP2 (Gaussian copula with log-normal marginals); $\sigma_j$ is floored at 0.05}
\label{tbl:dgp2}
\begin{tabular*}{\textwidth}{@{\extracolsep\fill}ll@{}}
\toprule
Parameter & Expression \\
\midrule
$\mu_1(x)$ & $1.20 + 0.80\,x_6 + 0.60\,x_1 - 0.50\,x_0 - 0.40\,x_2 x_4 + 0.30\,x_3 x_6$ \\
$\mu_2(x)$ & $1.10 + 0.70\,x_2 + 0.50\,x_4 - 0.45\,x_0 + 0.35\,x_1^2 - 0.25\,x_6^2$ \\
$\mu_3(x)$ & $1.30 + 0.90\,x_3 - 0.60\,x_0 + 0.40\,x_5 x_6 + 0.30\,x_1 x_3 - 0.20\,x_4^2$ \\
$\mu_4(x)$ & $1.00 + 0.65\,x_4 + 0.45\,x_6 - 0.35\,x_0 + 0.25\,x_2 x_5 - 0.15\,x_1 x_6$ \\
$\mu_5(x)$ & $0.90 + 0.55\,x_5 + 0.40\,x_1 - 0.30\,x_0 + 0.20\,x_3 x_4 - 0.12\,x_2^2$ \\
\midrule
$\sigma_1(x)$ & $0.25 + 0.50\,x_6(1-x_1) + 0.20\,x_0$ \\
$\sigma_2(x)$ & $0.22 + 0.45(1-x_4) + 0.18\,x_6 + 0.15\,x_0$ \\
$\sigma_3(x)$ & $0.20 + 0.40(1-x_3) + 0.15\,x_6 + 0.12\,x_0$ \\
$\sigma_4(x)$ & $0.18 + 0.35(1-x_5) + 0.12\,x_6 + 0.10\,x_0$ \\
$\sigma_5(x)$ & $0.15 + 0.30(1-x_2) + 0.10\,x_6 + 0.08\,x_0$ \\
\midrule
$R_{12}, R_{13}$ & $0.45 + 0.30\,x_6$, \quad $0.25 + 0.25(1-x_1)$ \\
$R_{14}, R_{15}$ & $0.20 + 0.20(1-x_2)$, \quad $0.15 + 0.15(1-x_3)$ \\
$R_{23}, R_{24}$ & $0.40 + 0.30(1-2x_4)$, \quad $0.25 + 0.20(1-x_5)$ \\
$R_{25}, R_{34}$ & $0.20 + 0.15(1-x_6)$, \quad $0.30 + 0.25(1-x_3)$ \\
$R_{35}, R_{45}$ & $0.22 + 0.18(1-x_4)$, \quad $0.35 + 0.28(1-2x_5)$ \\
\botrule
\end{tabular*}
\end{table}

\section{External problems}\label{app:external}\setcounter{table}{0}

All decisions are scaled to $[0,1]$ unless stated otherwise, $L_\varrho$ denotes the Cholesky factor of the equicorrelation matrix with parameter $\varrho$, and the loss is minimised.

In E1 (pricing and ordering), the prices are $p=5+20x_{1:2}$ and the order quantities $Q=80x_{3:4}$. Demand $D_j$ is gamma with mean $\bar D_1=60\exp\{-0.18(p_1-5)+0.06(p_2-5)\}$, $\bar D_2=45\exp\{-0.15(p_2-5)+0.05(p_1-5)\}$ and shape $2+0.3(p_j-5)$, and the two demands are coupled by a Gaussian copula with correlation 0.4. The loss is minus the profit $\sum_j[p_j\min(Q_j,D_j)-c_jQ_j+(Q_j-D_j)_+-2(D_j-Q_j)_+]$ with $c=(4,3.5)$.

In E2 (portfolio with market impact and crowding), the weights are $x_j\in[0,0.4]$ with $\sum_jx_j\le1$ (rescaled if violated); the rest earns $r_f=0.02$. Returns are $R=\bar r-\kappa\odot x+\sigma(x)\odot(L_{0.3}Z_{1:5})$ with $\sigma_j(x)=\varsigma_j(1+0.5x_j)$, $\bar r=(0.08,0.10,0.12,0.14,0.16)$, $\varsigma=(0.05,0.07,0.09,0.11,0.13)$ and $\kappa=(0.02,0.03,0.04,0.06,0.08)$. With probability $0.03+0.3\|x\|^2$ (driven by $Z_6$) a crash occurs, in which the mean is lowered by $0.15\,b$ with $b=(0.6,0.8,1.0,1.2,1.4)$ and the volatility is multiplied by 1.5. The loss is $-(x^\top R+(1-\mathbf{1}^\top x)r_f)$.

In E3 (service system with induced demand), service rates $\mu_j=1+2x_j$ attract arrival rates $\lambda_j=\lambda^0_j\{1+0.6(1-\mathrm{e}^{-0.8(\mu_j-1)})\}$ with $\lambda^0=(0.6,0.7,0.8)$. The waiting time at station $j$ is zero with probability $1-\lambda_j/\mu_j$ and otherwise exponential with rate $\mu_j-\lambda_j$. The loss is $\sum_j(c_j\mu_j-5\lambda_j)+\sum_j4\lambda_jW_j$ with $c=(3,3.5,4)$.

In E4 (project crashing), five activities have log-normal durations with means $m^0_j(1-0.4x_j)$, $m^0=(4,6,5,3,7)$, and coefficients of variation $0.25(1+0.5x_j)$, coupled through $L_{0.3}$ in log space. The makespan is $\max\{W_1+W_2,\,W_1+W_3+W_4,\,W_5+W_4\}$ and the loss is $\sum_j1.5c_jx_j^2+15(\text{makespan}-11)_+$ with $c=(8,10,6,9,5)$.

In E5 (insurance deductibles), the deductibles are $k_j=0.5+4.5x_j$ on four lines. Claims arrive at rate $\lambda_j=\lambda^0_j\,\mathrm{e}^{-0.15k_j}$ with $\lambda^0=(40,30,20,10)$ and have log-normal severities with log-location $(1.0,1.3,1.6,2.0)$ and log-scale $(0.8,0.9,1.0,1.2)$. The paid loss of each line is log-normal with the mean and variance of the compound Poisson sum of $(Y-k_j)_+$, coupled through $L_{0.5}$; with probability 0.02 (driven by $Z_5$) all paid losses are tripled. The loss is $\sum_jW_j-\sum_jp^0_j(1-0.15k_j)$ with $p^0=(150,140,120,100)$.

In E6 (preventive maintenance), the failure times of three machines are Weibull with scale $\epsilon^0_j(1+1.5x_j)$, $\epsilon^0=(6,8,7)$, and shape $1.2+1.5x_j$, coupled by a Gaussian copula with correlation 0.3. Over a horizon of 10, the loss is $\sum_j10\,c^m_jx_j+\sum_j10\,c^r_j(10-W_j)_+$ with $c^m=(9,12,15)$ and $c^r=(2,2.5,3)$.

\section{Budget-dependent settings}\label{app:settings}\setcounter{table}{0}

Table~\ref{tab:settings} lists the settings that depend on the budget. ACRS rounds $0.1B$ and $0.05B$ to the nearest integer; the reranking budget is spent only if more than one candidate remains. The refinement sizes are $n_\ell=\max\{80,\lfloor300\sqrt{B/260000}\rceil\}$ and $n_s=\max\{150,\lfloor500\sqrt{B/260000}\rceil\}$. SGD-CVaR runs $T=\lfloor0.85B/(40d)\rfloor$ iterations in total, with a quarter in each warm-up chain and the remainder in the main chain. Checkpoint estimates use 60 draws.

\begin{table}[!htb]
\caption{Budget-dependent settings. EI: expected improvement. CEM population: analytic and external problems; on DGP1/DGP2 the population is 30 at every budget}
\label{tab:settings}
\begin{tabular*}{\textwidth}{@{\extracolsep\fill}lrrrr@{}}
\toprule
Setting & $B=2000$ & $B=5000$ & $B=10000$ & $B=30000$ \\
\midrule
ACRS: design $n_0$ & 200 & 500 & 1000 & 3000 \\
ACRS: augmentation $n_a$ & 100 & 250 & 500 & 1500 \\
ACRS: reranking (at most) & 320 & 320 & 500 & 1500 \\
Refinement: sample-path size $n_\ell$ & 80 & 80 & 80 & 102 \\
Refinement: selection/pilot size $n_s$ & 150 & 150 & 150 & 170 \\
GP-BO: draws per evaluation & 50 & 50 & 50 & 115 \\
GP-BO: evaluations (EI steps) & 30 (0) & 75 (45) & 150 (120) & 195 (165) \\
CEM: draws per candidate & 10 & 10 & 11 & 33 \\
CEM: population & 10 & 16 & 30 & 30 \\
KDE-SO: design size & 400 & 1000 & 2000 & 2600 \\
\botrule
\end{tabular*}
\end{table}

\section{Per-setting results for DGP1 and DGP2}\label{app:tables}\setcounter{table}{0}

\begin{table}[!htb]
\caption{DGP1 and DGP2: median normalised regret over 30 replications. Bold: lowest median in the row. $^{+}$/$^{-}$: the competitor is significantly worse/better than ACRS (paired Wilcoxon test over replications, Holm-adjusted, level 0.05)}
\label{tab:dgp}
\begin{tabular*}{\textwidth}{@{\extracolsep\fill}lllrrrrrr@{}}
\toprule
DGP & $\lambda$ & $B$ (k) & ACRS & GP-BO & CEM & KDE-SO & SGD-CVaR & Refine-only \\
\midrule
DGP1 & 0.5 & 5 & \textbf{0.5} & 3.9$^{+}$ & 7.0$^{+}$ & 0.7$^{+}$ & 68.2$^{+}$ & 5.5$^{+}$ \\
DGP1 & 0.5 & 10 & 0.2 & 1.3$^{+}$ & \textbf{0.2} & 0.2 & 3.7$^{+}$ & 0.2 \\
DGP1 & 0.5 & 30 & 0.1 & 0.5 & 0.1 & 0.1 & 0.5 & \textbf{0.1} \\
DGP1 & 0.7 & 5 & 1.1 & 5.5$^{+}$ & 6.5$^{+}$ & \textbf{0.7} & 63.1$^{+}$ & 5.8$^{+}$ \\
DGP1 & 0.7 & 10 & 0.1 & 1.9$^{+}$ & 0.3 & 0.6$^{+}$ & 3.5$^{+}$ & \textbf{0.0} \\
DGP1 & 0.7 & 30 & 0.3 & 0.2 & \textbf{0.1} & 0.1 & 0.8 & 0.1 \\
DGP1 & 0.9 & 5 & \textbf{0.5} & 4.3$^{+}$ & 9.5$^{+}$ & 2.1$^{+}$ & 59.4$^{+}$ & 5.4$^{+}$ \\
DGP1 & 0.9 & 10 & \textbf{0.2} & 1.5$^{+}$ & 0.4 & 0.3 & 3.5$^{+}$ & 0.3 \\
DGP1 & 0.9 & 30 & 0.4 & 0.3 & 0.2 & \textbf{0.0}$^{-}$ & 0.4 & 0.2 \\
DGP2 & 0.5 & 5 & 0.1 & 1.7$^{+}$ & \textbf{0.1} & 1.0$^{+}$ & 37.2$^{+}$ & 0.3$^{+}$ \\
DGP2 & 0.5 & 10 & 0.1 & 0.1 & \textbf{0.0} & 0.1 & 1.6$^{+}$ & 0.1 \\
DGP2 & 0.5 & 30 & 0.0 & 0.0 & \textbf{0.0} & 0.0 & 0.2 & 0.1 \\
DGP2 & 0.7 & 5 & \textbf{0.1} & 0.8$^{+}$ & 0.1 & 0.6$^{+}$ & 34.6$^{+}$ & 0.3$^{+}$ \\
DGP2 & 0.7 & 10 & 0.0 & 0.1 & 0.0 & \textbf{0.0} & 0.8$^{+}$ & 0.1 \\
DGP2 & 0.7 & 30 & 0.0 & 0.1 & \textbf{0.0} & 0.0 & 0.0 & 0.2 \\
DGP2 & 0.9 & 5 & \textbf{0.0} & 0.9$^{+}$ & 0.1 & 0.1$^{+}$ & 42.5$^{+}$ & 0.4$^{+}$ \\
DGP2 & 0.9 & 10 & 0.1 & 0.1 & \textbf{0.0} & 0.1 & 0.5$^{+}$ & 0.3$^{+}$ \\
DGP2 & 0.9 & 30 & 0.1 & 0.1 & \textbf{0.0} & 0.0 & 0.0 & 0.1 \\
\botrule
\end{tabular*}
\end{table}

\end{appendices}

\end{document}